# Neural Network Optimal Power Flow via Energy Gradient Flow and Unified Dynamics

Liu Xuezhi

Shanghai Jiao Tong University

China

liuxz@sjtu.edu.cn

## Abstract

Optimal Power Flow (OPF) is a core optimization problem in power system operation and planning, aiming to minimize generation costs while satisfying physical constraints such as power flow equations, generator limits, and voltage limits. Traditional OPF solving methods typically employ iterative optimization algorithms (such as interior point methods, sequential quadratic programming, etc.), with limitations including low computational efficiency, initial value sensitivity, and low batch computation efficiency. Most existing deep learning-based OPF methods rely on supervised learning, requiring pre-solving large numbers of cases, and have difficulty guaranteeing physical consistency.

This paper proposes an Optimal Power Flow solving method based on neural network dynamics and energy gradient flow, transforming OPF problems into energy minimization problems. By constructing an energy function $V(\mathbf{x}, \mathbf{P}_g) = \frac{1}{2}\|\mathbf{F}(\mathbf{x}, \mathbf{P}_g)\|^2$ to measure the degree of deviation from the constraint manifold, and guiding networks to learn optimal solutions that simultaneously satisfy power flow constraints and minimize costs through gradient flow $\frac{d(\mathbf{x}, \mathbf{P}_g)}{dt} = -\nabla V(\mathbf{x}, \mathbf{P}_g)$. Neural networks are trained unsupervised by directly minimizing physical residuals, requiring no labeled data, achieving true "end-to-end" physics-constrained learning.

The core innovations of this method include: (1) **Unified Dynamics Framework**: Treating the cost function as "economic potential," unifying it with the physical energy function into a total energy function, dynamically balanced through Lagrangian multiplier $\lambda_F$ (shadow price), avoiding gradient competition between cost terms and energy terms in traditional methods; (2) **Energy Landscape Synthesis**: Adopting augmented Lagrangian method to construct unified energy landscape, automatically adjusting coupling strength between economic and physical potentials through dynamic Lagrangian multipliers, enabling intelligent balancing at constraint boundaries; (3) **Architecture-Level Physical Embedding**: Treating voltage limits as "boundary conditions," automatically satisfied through network architecture design (tanh mapping), requiring no penalty terms in the loss function; (4) **Multi-Component Loss Function**: Treating different constraints as "guiding signals" or "hard constraints," using different strategies such as ReLU functions and exponential penalties, forming flexible multi-component loss function design; (5) **Breakthrough Path Strategy**: Through three key improvements—dynamic weighting, warm-up restart, and physics-guided sampling—achieving breakthrough from "local exploitation" to "global exploration," avoiding getting stuck in local optima.

# 1. Introduction

## 1.1 Problem Background and Challenges

Optimal Power Flow (OPF) is a core optimization problem in power system operation and planning, aiming to minimize generation costs while satisfying physical constraints such as power flow equations, generator output limits, and voltage limits. The OPF problem can be formulated as:

$$\begin{aligned}
\min_{\mathbf{x},\mathbf{P}_g} \quad & C(\mathbf{P}_g) = \sum_{i=1}^{n_g} C_i(P_{g,i}) \\
\text{s.t.} \quad & \mathbf{F}(\mathbf{x}, \mathbf{P}_g) = \mathbf{0} \quad \text{(Power Flow Equations)} \\
& \mathbf{P}_g^{\min} \leq \mathbf{P}_g \leq \mathbf{P}_g^{\max} \quad \text{(Generator Limits)} \\
& \mathbf{V}^{\min} \leq \mathbf{V} \leq \mathbf{V}^{\max} \quad \text{(Voltage Limits)}
\end{aligned} \quad (1)$$

where $\mathbf{x} = [V_1, \ldots, V_n, \theta_1, \ldots, \theta_n]^\top$ is the system state vector (voltage magnitudes and phase angles), $\mathbf{P}_g = [P_{g,1}, \ldots, P_{g,n_g}]^\top$ is the generator active power output vector, $C(\mathbf{P}_g)$ is the total generation cost function, and $\mathbf{F}(\mathbf{x}, \mathbf{P}_g) = \mathbf{0}$ is the power flow equation constraint.

Traditional OPF solving methods typically employ iterative optimization algorithms (such as interior point methods [16], sequential quadratic programming [17], etc.), with the following fundamental limitations:

**Computational Efficiency**: Each solution requires iterative optimization, with computation time growing superlinearly with system scale. When processing large numbers of scenarios (e.g., Monte Carlo analysis, inner-layer solving in optimization problems), each scenario requires independent optimization, with computational cost growing linearly with the number of scenarios.

**Initial Value Sensitivity**: Requires appropriate initial guesses, otherwise may converge to local optima or diverge. For large-scale systems or heavy-load scenarios, finding appropriate initial values is itself a difficult problem.

**Low Batch Computation Efficiency**: Traditional methods have difficulty fully utilizing parallel computing advantages, with computational efficiency in batch scenarios significantly lower than neural network methods.

## 1.2 Limitations of Existing Machine Learning Methods

In recent years, deep learning has shown great potential in scientific computing, but for the specific problem of OPF solving, existing methods still have obvious defects:

**Problems with Supervised Learning Paradigm**: Most neural network-based OPF solvers adopt supervised learning frameworks, i.e., pre-solving large numbers of OPF cases using traditional methods, constructing $(\text{load condition}, \text{OPF solution})$ datasets, then training networks to learn this mapping [18,19,20]. These methods have three fundamental problems: (1) **Strong data dependency**: Require pre-solving large numbers of cases, with high computational cost; (2) **Limited generalization ability**: Networks can only learn patterns within the training data distribution, lacking robustness to unseen load combinations or topology changes; (3) **No guarantee of physical consistency**: Network outputs may not strictly satisfy power flow equations, requiring post-processing or additional constraints.

**Limitations of PINN Methods**: Physics-Informed Neural Networks (PINN) attempt to balance observational data and physical laws by including both data fitting terms and physical constraint terms in the loss function [6,7]. However, in OPF problems, PINN faces unique challenges: (1) **Missing boundary conditions**: Unlike PDE solving, OPF is a constrained optimization problem lacking explicit boundary/initial conditions; (2) **Multi-objective balancing**: Need to simultaneously optimize cost and satisfy constraints, making PINN's weight balancing problem more complex; (3) **Constraint handling difficulty**: Inequality constraints (e.g., generator limits, voltage limits) are difficult to handle naturally through the PINN framework.

**Lack of Unified Theoretical Framework**: Most existing methods treat OPF solving as a "black-box" optimization problem, ignoring the geometric structure (constraint manifold) and dynamical properties (gradient flow) of the problem. This lack of perspective leads to methods lacking interpretability and difficulty establishing theoretical connections with classical optimization methods.

## 1.3 Core Motivation of This Method

Based on the **Physics-Constrained Neural Dynamics Framework** and energy gradient flow, we propose an end-to-end OPF solving method, aiming to solve the following core problems:

**Label-Free Learning**: By directly minimizing power flow equation residuals $\|\mathbf{F}(\mathbf{x}, \mathbf{P}_g)\|^2$, avoid dependence on pre-solved data, achieving true "end-to-end" physics-constrained learning. This completely eliminates dependence on labeled data, significantly reducing computational cost.

**Unified Dynamics Framework**: Treat the cost function as "economic potential," unifying it with the physical energy function into a total energy function. Through dynamic balancing via Lagrangian multiplier $\lambda_F$ (shadow price), avoid gradient competition between cost terms and energy terms in traditional methods, achieving theoretical unification and training stability improvement.

**Architecture-Level Physical Embedding**: Treat voltage limits as "boundary conditions," automatically satisfied through network architecture design (tanh mapping), requiring no penalty terms in the loss function. This is a typical application of architecture-level constraint satisfaction, improving training stability and computational efficiency.

**Energy Landscape Synthesis**: Adopt augmented Lagrangian method to construct unified energy landscape, automatically adjusting coupling strength between economic and physical potentials through dynamic Lagrangian multipliers. Lagrangian multiplier $\lambda_F$ has clear cost semantics (shadow price), representing marginal cost at the optimal solution, giving the optimization process clear physical interpretation.

**Breakthrough Path Strategy**: Through three key improvements—dynamic weighting, warm-up restart, and physics-guided sampling—achieve breakthrough from "local exploitation" to "global exploration," avoid getting stuck in local optima, and improve global optimization capability.

## 1.4 Core Contributions

The core contributions of this paper are:

1. **Unified Dynamics Framework**: For the first time, unify cost optimization and constraint satisfaction under one energy framework. Through unification of "economic potential" and "physics potential," achieve theoretical unification and training stability improvement. Through dynamic balancing of economic potential and physics potential via Lagrangian multiplier $\lambda_F$ (shadow price), with clear cost semantics, no manual weight tuning required.

2. **Energy Landscape Synthesis Framework**: Adopt augmented Lagrangian method to construct unified energy landscape, automatically adjusting coupling strength between economic and physics potentials through dynamic Lagrangian multipliers (dual ascent method). Lagrangian multiplier $\lambda_F$ has clear cost semantics (shadow price), satisfying KKT conditions at the optimal solution, giving the optimization process clear physical interpretation.

3. **Architecture-Level Physical Embedding**: Treat voltage limits as "boundary conditions," automatically satisfied through network architecture design (tanh mapping), requiring no penalty terms in the loss function. This design improves training stability, reduces constraint violations, and simplifies loss function design.

4. **Multi-Component Loss Function**: Treat different constraints as "guiding signals" or "hard constraints,"

using different strategies such as ReLU functions and exponential penalties, forming flexible multi-component loss function design. This design allows networks to gradually learn to satisfy constraints, making training more stable.
   5. **Breakthrough Path Strategy**: Through three key improvements—dynamic weighting, warm-up restart (Best Loss stagnation detection), and physics-guided sampling—achieve breakthrough from "local exploitation" to "global exploration." These three improvements work together, forming a complete global optimization breakthrough path.
   6. **Label-Free Learning Paradigm**: Directly minimize power flow equation residuals, requiring no pre-solved data, achieving true "end-to-end" physics-constrained learning. This completely eliminates dependence on labeled data, significantly reducing computational cost.

## 1.5 Differences from Power Flow (PF) Problems

It is important to emphasize that OPF problems are fundamentally different from Power Flow (PF) problems:

**PF Problem**: Given generator outputs and loads, solve for system state (voltages, phase angles) such that power flow equations are satisfied. This is a **constraint satisfaction problem**, with the goal of finding solutions that satisfy physical constraints.

**OPF Problem**: While satisfying physical constraints, optimize generator outputs to minimize generation cost. This is a **constrained optimization problem**, with the goal of finding solutions that both satisfy physical constraints and are economically optimal.

**Key Differences**:

- **Increased Degrees of Freedom**: In OPF problems, generator outputs $\mathbf{P}_g$ change from fixed parameters to decision variables, increasing degrees of freedom by $n_g$ (number of generators)
- **Objective Function**: OPF problems introduce a cost function $C(\mathbf{P}_g)$, requiring cost minimization while satisfying constraints
- **Constraint Manifold Extension**: The constraint manifold expands from $\mathcal{M}_{\mathrm{PF}} = \{\mathbf{x} \mid \mathbf{F}(\mathbf{x}, \mathbf{P}_g^{\mathrm{fixed}}) = \mathbf{0}\}$ to $\mathcal{M}_{\mathrm{OPF}} = \{(\mathbf{x}, \mathbf{P}_g) \mid \mathbf{F}(\mathbf{x}, \mathbf{P}_g) = \mathbf{0}, \mathbf{P}_g^{\min} \leq \mathbf{P}_g \leq \mathbf{P}_g^{\max}\}$

This method extends the physics-constrained neural dynamics framework from PF problems to OPF problems, achieving unification of cost optimization and constraint satisfaction through unified dynamics framework and energy landscape synthesis.

# 2. Related Work

## 2.1 Traditional OPF Solving Methods

Since the 1960s, Interior Point Methods and their variants have been the mainstream for OPF solving [16,17]. These methods transform constrained optimization problems into unconstrained problems by converting inequality constraints into logarithmic barrier functions, then solve using Newton's method. Although they perform well in most scenarios, traditional methods have the following fundamental limitations:

**Computational Efficiency**: Each solution requires iterative optimization, with computation time growing superlinearly with system scale. For large-scale systems (e.g., IEEE 300+ buses), a single solution may take seconds or even minutes.

**Initial Value Sensitivity**: Requires appropriate initial guesses, otherwise may converge to local optima or diverge. For large-scale systems or heavy-load scenarios, finding appropriate initial values is itself a difficult problem.

**Low Batch Computation Efficiency**: When processing large numbers of scenarios (e.g., Monte Carlo analysis, inner-layer solving in optimization problems), each scenario requires independent optimization, with computational cost growing linearly with the number of scenarios.

**Jacobian Matrix Computation Overhead**: Each iteration requires computing and decomposing the Jacobian matrix. For large-scale systems, the computational and storage costs of this step are significant.

## 2.2 Machine Learning-Based OPF Solving

In recent years, deep learning has shown potential in OPF solving. **Supervised learning methods** pre-solve large numbers of OPF cases using traditional methods, construct $(\text{load condition}, \text{OPF solution})$ datasets, and train neural networks to learn this mapping [18,19,20]. These methods can achieve high accuracy in specific scenarios, but have three fundamental problems: (1) **Strong data dependency**: Requires pre-solving large numbers of cases, with high computational cost; (2) **Limited generalization**: Networks can only learn patterns within the training data distribution, lacking robustness to unseen load combinations or topology changes; (3) **No guarantee of physical consistency**: Network outputs may not strictly satisfy power flow equations, requiring post-processing or additional constraints.

**Reinforcement learning methods** model OPF solving as a sequential decision problem, learning solving strategies through interaction with the environment. These methods have certain advantages in dynamic scenarios, but the training process is complex and it is difficult to guarantee physical consistency of solutions.

## 2.3 Physics-Informed Neural Networks (PINN) in Optimization Problems

Physics-Informed Neural Networks (PINN) attempt to balance observational data and physical laws by including both data fitting terms and physical constraint terms in the loss function [6,7]. PINN has achieved significant success in PDE solving, parameter identification, and other fields. However, in OPF, a constrained optimization problem, PINN faces unique challenges:

**Missing Boundary Conditions**: Unlike PDE solving, OPF is a constrained optimization problem lacking explicit boundary/initial conditions, making it difficult to define PINN's "data term."

**Multi-objective Balancing**: Need to simultaneously optimize cost and satisfy constraints, making PINN's weight balancing problem more complex. Traditional PINN methods require manual adjustment of multiple weight coefficients, making tuning difficult.

**Constraint Handling Difficulty**: Inequality constraints (e.g., generator limits, voltage limits) are difficult to handle naturally through the PINN framework. Existing methods typically use penalty terms, but require careful tuning of penalty coefficients, otherwise may lead to constraint violations or insufficient optimization.

## 2.4 Manifold Geometry and Constrained Optimization

Manifold optimization methods treat constrained optimization problems as unconstrained optimization on manifolds, solving through geometric operations such as tangent space projection and retraction mapping [8,9]. These methods have good convergence guarantees in theory, but in practice, explicit construction and maintenance of manifold structures often incur significant computational overhead.

In OPF problems, the constraint manifold is defined as:

$$\mathcal{M}_{\text{OPF}} = \left\{ (\mathbf{x}, \mathbf{P}_g) \in \mathbb{R}^{2n+n_g} \mid \mathbf{F}(\mathbf{x}, \mathbf{P}_g) = \mathbf{0}, \mathbf{P}_g^{\min} \leq \mathbf{P}_g \leq \mathbf{P}_g^{\max} \right\} \qquad (2)$$

This method is the first to combine manifold geometry perspective with neural networks, providing structured learning objectives for neural networks through explicit construction of constraint manifolds and utilization of tangent and normal space structures. At the same time, the energy function $V(\mathbf{x}, \mathbf{P}_g) = \frac{1}{2}\|\mathbf{F}(\mathbf{x}, \mathbf{P}_g)\|^2$ measures the degree of deviation from the constraint manifold, enabling the network to learn to "slide into" the constraint manifold.

## 2.5 Augmented Lagrangian Method in Neural Network Optimization

The Augmented Lagrangian Method is a classical method for handling constrained optimization problems, transforming constrained optimization problems into unconstrained problems by introducing Lagrangian multipliers and penalty terms [21,22]. In recent years, the augmented Lagrangian method has been widely applied in neural network optimization, particularly in Generative Adversarial Networks (GAN) and Variational Autoencoders (VAE) [23,24].

This method innovatively applies the augmented Lagrangian method to neural network solving of OPF problems, automatically adjusting the coupling strength between economic and physical potentials through dynamic Lagrangian multipliers. The Lagrangian multiplier $\lambda_F$ has clear cost semantics (shadow price), representing marginal cost at the optimal solution, giving the optimization process clear physical interpretation.

## 2.6 Differences Between This Method and Existing Methods

Compared to **supervised learning methods**, this method completely eliminates dependence on labeled data, achieving true "unsupervised physics-constrained learning." Networks are trained by directly minimizing power flow equation residuals, without pre-solving any cases, significantly reducing computational cost.

Compared to **PINN methods**, this method adopts an augmented Lagrangian framework, automatically balancing cost and constraints through dynamic Lagrangian multipliers, avoiding weight tuning problems. More importantly, this method treats voltage limits as "boundary conditions," automatically satisfied through network architecture design, without additional penalty terms.

Compared to **traditional manifold optimization methods**, this method learns mappings on manifolds through neural networks, avoiding computational overhead of explicit construction and maintenance of manifold structures. At the same time, the parallel computing capability of neural networks significantly improves computational efficiency in batch scenarios.

Compared to **traditional augmented Lagrangian methods**, this method closely integrates Lagrangian multipliers with the neural network training process, automatically updating multipliers through gradient ascent, without manual adjustment. The Lagrangian multiplier $\lambda_F$ has clear cost semantics (shadow price), satisfying KKT conditions at the optimal solution, giving the optimization process clear physical interpretation.

Overall, this method organically combines **manifold geometry, gradient flow dynamics, neural networks, and augmented Lagrangian methods**, forming a unified, scalable, physically consistent OPF solving framework, providing new ideas for real-time optimization of large-scale power systems.

# 3. Problem Modeling

## 3.1 Mathematical Formulation of Optimal Power Flow Problem

Consider a power system with $n$ buses, containing $n_g$ generators. The Optimal Power Flow (OPF) problem aims to minimize total generation cost while satisfying physical constraints. This problem can be formulated as the following constrained optimization problem:

$$\begin{aligned}
\min_{\mathbf{x},\mathbf{P}_g} \quad & C(\mathbf{P}_g) = \sum_{i=1}^{n_g} C_i(P_{g,i}) \\
\text{s.t.} \quad & \mathbf{F}(\mathbf{x}, \mathbf{P}_g) = \mathbf{0} \quad \text{(Power Flow Equations)} \\
& \mathbf{P}_g^{\min} \leq \mathbf{P}_g \leq \mathbf{P}_g^{\max} \quad \text{(Generator Output Limits)} \\
& \mathbf{V}^{\min} \leq \mathbf{V} \leq \mathbf{V}^{\max} \quad \text{(Voltage Limits)} \\
& \mathbf{Q}_g^{\min} \leq \mathbf{Q}_g \leq \mathbf{Q}_g^{\max} \quad \text{(Reactive Power Limits)}
\end{aligned} \quad (3)$$

where:

- $\mathbf{x} = [V_1, \ldots, V_n, \theta_1, \ldots, \theta_n]^\top \in \mathbb{R}^{2n}$ is the system state vector, containing voltage magnitudes $V_i$ and phase angles $\theta_i$ for all buses
- $\mathbf{P}_g = [P_{g,1}, \ldots, P_{g,n_g}]^\top \in \mathbb{R}^{n_g}$ is the generator active power output vector (decision variables)
- $C(\mathbf{P}_g) = \sum_{i=1}^{n_g} C_i(P_{g,i})$ is the total generation cost function, where $C_i(P_{g,i})$ is typically a quadratic function: $C_i(P_{g,i}) = a_i P_{g,i}^2 + b_i P_{g,i} + c_i$
- $\mathbf{F}(\mathbf{x}, \mathbf{P}_g) = \mathbf{0}$ is the power flow equation constraint, describing active and reactive power balance at each bus

### 3.1.1 Detailed Form of Power Flow Equations

For bus $i$, the power flow equations describe active and reactive power balance:

$$\begin{aligned}
P_i &= V_i \sum_{j=1}^{n} V_j \left[ G_{ij} \cos(\theta_i - \theta_j) + B_{ij} \sin(\theta_i - \theta_j) \right] \\
Q_i &= V_i \sum_{j=1}^{n} V_j \left[ G_{ij} \sin(\theta_i - \theta_j) - B_{ij} \cos(\theta_i - \theta_j) \right]
\end{aligned} \quad (4)$$

where $G_{ij}$ and $B_{ij}$ are the real and imaginary parts of the bus admittance matrix $\mathbf{Y} = \mathbf{G} + j\mathbf{B}$. For buses with generators, the net injection power is $P_i = P_{g,i} - P_{d,i}$ (generation minus load), and for pure load buses, $P_i = -P_{d,i}$.

Combining equations for all buses, we obtain a nonlinear system of equations:

$$\mathbf{F}(\mathbf{x}, \mathbf{P}_g) = \begin{bmatrix} \mathbf{P}_{\text{spec}} - \mathbf{P}_{\text{calc}}(\mathbf{x}, \mathbf{P}_g) \\ \mathbf{Q}_{\text{spec}} - \mathbf{Q}_{\text{calc}}(\mathbf{x}) \end{bmatrix} = \mathbf{0} \quad (5)$$

where $\mathbf{P}_{\text{spec}}$ and $\mathbf{Q}_{\text{spec}}$ are the specified power injection vectors (depending on $\mathbf{P}_g$ and loads), and $\mathbf{P}_{\text{calc}}$ and $\mathbf{Q}_{\text{calc}}$ are the power injections calculated from the state vector $\mathbf{x}$.

### 3.1.2 Bus Types and Constraint Structure

Buses in power systems are classified into three types based on their physical characteristics:

1. **SLACK Bus (Reference Bus)**: Voltage magnitude $V$ and phase angle $\theta$ are fixed, typically as the reference bus ($\theta = 0$). The generator output $P_{g,\text{slack}}$ at this bus is not a decision variable, but is automatically determined by power balance.
2. **PV Bus (Voltage-Controlled Bus)**: Voltage magnitude $V$ is fixed, active power $P_g$ is a decision variable, and reactive power $Q_g$ is determined by power flow equations (must satisfy $Q_g^{\min} \leq Q_g \leq Q_g^{\max}$).
3. **PQ Bus (Load Bus)**: Active and reactive powers $P_d, Q_d$ are fixed, voltage magnitude $V$ and phase angle $\theta$ are state variables, determined by power flow equations.

In OPF problems, **the voltage $V$ of PV buses can also be a decision variable** (optimized within $[V^{\min}, V^{\max}]$ range), which increases the problem's degrees of freedom but also expands the optimization space.

## 3.2 Geometric Structure of Constraint Manifold

The power flow equations $\mathbf{F}(\mathbf{x}, \mathbf{P}_g) = \mathbf{0}$ define a constraint manifold embedded in the $(\mathbf{x}, \mathbf{P}_g)$ joint space:

$$\mathcal{M}_{\text{OPF}} = \left\{ (\mathbf{x}, \mathbf{P}_g) \in \mathbb{R}^{2n+n_g} \mid \mathbf{F}(\mathbf{x}, \mathbf{P}_g) = \mathbf{0}, \mathbf{P}_g^{\min} \leq \mathbf{P}_g \leq \mathbf{P}_g^{\max} \right\} \tag{6}$$

This manifold has the following important geometric properties:

### 3.2.1 Tangent Space and Normal Space

At a point $(\mathbf{x}, \mathbf{P}_g) \in \mathcal{M}_{\text{OPF}}$ on the manifold, the Jacobian matrix $\mathbf{J}(\mathbf{x}, \mathbf{P}_g) = \frac{\partial \mathbf{F}}{\partial (\mathbf{x}, \mathbf{P}_g)}$ defines the tangent space and normal space:

**Tangent Space**:

$$T_{(\mathbf{x}, \mathbf{P}_g)} \mathcal{M}_{\text{OPF}} = \left\{ \mathbf{v} \in \mathbb{R}^{2n+n_g} \mid \mathbf{J}(\mathbf{x}, \mathbf{P}_g) \mathbf{v} = \mathbf{0} \right\} \tag{7}$$

The tangent space describes locally feasible directions on the manifold. Search directions of traditional optimization methods (e.g., interior point methods) lie in the tangent space.

**Normal Space**:

$$N_{(\mathbf{x}, \mathbf{P}_g)} \mathcal{M}_{\text{OPF}} = \left\{ \mathbf{w} \in \mathbb{R}^{2n+n_g} \mid \mathbf{w} = \mathbf{J}(\mathbf{x}, \mathbf{P}_g)^\top \boldsymbol{\lambda}, \boldsymbol{\lambda} \in \mathbb{R}^m \right\} \tag{8}$$

The normal space describes directions deviating from the manifold, where $m$ is the dimension of power flow equations. The gradient of the energy function $\nabla V(\mathbf{x}, \mathbf{P}_g)$ lies in the normal space, pointing in the direction of fastest energy increase.

### 3.2.2 Manifold Extension from PF to OPF

It is important to emphasize that OPF problems are fundamentally different from Power Flow (PF) problems:

**PF Problem**: Given generator outputs $\mathbf{P}_g^{\text{fixed}}$ and loads, solve for system state $\mathbf{x}$ such that power flow equations are satisfied. This is a **constraint satisfaction problem**, with constraint manifold:

$$\mathcal{M}_{\text{PF}} = \left\{ \mathbf{x} \in \mathbb{R}^{2n} \mid \mathbf{F}(\mathbf{x}, \mathbf{P}_g^{\text{fixed}}) = \mathbf{0} \right\} \tag{9}$$

**OPF Problem**: While satisfying physical constraints, optimize generator outputs $\mathbf{P}_g$ and system state $\mathbf{x}$ to minimize generation cost. This is a **constrained optimization problem**, with constraint manifold extended to:

$$\mathcal{M}_{\text{OPF}} = \left\{ (\mathbf{x}, \mathbf{P}_g) \in \mathbb{R}^{2n+n_g} \mid \mathbf{F}(\mathbf{x}, \mathbf{P}_g) = \mathbf{0}, \mathbf{P}_g^{\min} \leq \mathbf{P}_g \leq \mathbf{P}_g^{\max} \right\} \tag{10}$$

**Key Differences**:

1. **Increased Degrees of Freedom**: In OPF problems, generator outputs $\mathbf{P}_g$ change from fixed parameters to decision variables, increasing degrees of freedom by $n_g$ (number of generators)
2. **Objective Function**: OPF problems introduce a cost function $C(\mathbf{P}_g)$, requiring cost minimization while satisfying constraints
3. **Manifold Dimension**: The constraint manifold expands from $2n - m$ dimensions to $2n + n_g - m$ dimensions (where $m$ is the dimension of power flow equations)

## 3.3 Energy Function and Gradient Flow

To measure the degree of deviation from the constraint manifold, we construct an energy function:

$$V(\mathbf{x}, \mathbf{P}_g) = \frac{1}{2}\|\mathbf{F}(\mathbf{x}, \mathbf{P}_g)\|_2^2 = \frac{1}{2}\left(\|\mathbf{P}_{\text{residual}}\|_2^2 + \|\mathbf{Q}_{\text{residual}}\|_2^2\right) \tag{11}$$

where $\mathbf{P}_{\text{residual}}$ and $\mathbf{Q}_{\text{residual}}$ are the active and reactive power residual vectors, respectively.

Physical meaning of the energy function:

- $V = 0$: Solution strictly satisfies power flow equations, located on the constraint manifold
- $V > 0$: Solution deviates from the constraint manifold; the larger the energy function value, the greater the deviation

The gradient of the energy function is:

$$\nabla V(\mathbf{x}, \mathbf{P}_g) = \mathbf{J}(\mathbf{x}, \mathbf{P}_g)^\top \mathbf{F}(\mathbf{x}, \mathbf{P}_g) \tag{12}$$

The gradient flow is defined as:

$$\frac{d(\mathbf{x}, \mathbf{P}_g)}{dt} = -\nabla V(\mathbf{x}, \mathbf{P}_g) = -\mathbf{J}(\mathbf{x}, \mathbf{P}_g)^\top \mathbf{F}(\mathbf{x}, \mathbf{P}_g) \tag{13}$$

This dynamical system describes an energy decay process: trajectories follow the direction of fastest energy decrease, eventually converging to the constraint manifold (energy minimum point, i.e., $V = 0$).

## 3.4 Unified Dynamics Framework: Economic Potential and Physical Potential

Traditional OPF solving methods typically treat cost optimization and constraint satisfaction as two independent objectives, coordinating through weight balancing or penalty terms. The core innovation of this method is to treat the cost function as "economic potential," unifying it with the physical energy function into a total energy function.

### 3.4.1 Economic Potential

Convert the cost function to "economic potential":

$$\Phi_{\text{economic}}(\mathbf{P}_g) = \left(\frac{C(\mathbf{P}_g)}{C_{\text{ref}}}\right)^2 \tag{14}$$

where $C_{\text{ref}}$ is a reference cost (for normalization, typically $C_{\text{ref}} = 10^4$ $/h). Economic potential quantifies the degree of deviation from optimal cost.

### 3.4.2 Unified Total Energy Function

Unify economic potential and physical potential into a total energy function:

$$E_{\text{total}}(\mathbf{x}, \mathbf{P}_g, \lambda_F) = \Phi_{\text{economic}}(\mathbf{P}_g) + \lambda_F \cdot V(\mathbf{x}, \mathbf{P}_g) \tag{15}$$

where $\lambda_F \geq 0$ is a Lagrangian multiplier (shadow price), dynamically updated through dual ascent method, with clear cost semantics. The unified energy function unifies cost optimization and constraint satisfaction under one framework, avoiding gradient competition between cost terms and energy terms in traditional methods.

### 3.4.3 Energy Landscape Synthesis: Augmented Lagrangian Framework

To more precisely balance economic optimization and physical constraints, we adopt the augmented Lagrangian method to construct a unified energy landscape:

$$\mathcal{L}(\mathbf{x}, \mathbf{P}_g, \lambda_F, \boldsymbol{\lambda}, \boldsymbol{\mu}) = C(\mathbf{P}_g) + \lambda_F \cdot V(\mathbf{x}, \mathbf{P}_g) + \sum_c \left(\lambda_c h_c + \frac{\mu_c}{2} h_c^2\right) \quad (16)$$

where:

- $C(\mathbf{P}_g)$ is the economic potential (objective function)
- $\lambda_F \cdot V(\mathbf{x}, \mathbf{P}_g)$ is the physical potential (constraint function), where $\lambda_F$ is the Lagrangian multiplier (shadow price)
- $\sum_c \left(\lambda_c h_c + \frac{\mu_c}{2} h_c^2\right)$ is the constraint potential (augmented Lagrangian terms), where $h_c$ is the constraint violation

The Lagrangian multiplier $\lambda_F$ has clear cost semantics (shadow price), representing marginal cost at the optimal solution, satisfying KKT conditions:

$$\lambda_F = -\frac{\partial C/\partial V}{\partial V/\partial \mathbf{P}_g} \quad \text{(at optimal solution)} \quad (17)$$

## 3.5 Neural Network Dynamics Perspective

Under the neural network dynamics framework, we let the neural network learn a mapping:

$$(\mathbf{x}_\theta(\mathbf{u}), \mathbf{P}_{g,\theta}(\mathbf{u})) = f_\theta(\mathbf{u}) \quad (18)$$

where $\mathbf{u} \in \mathbb{R}^d$ is the load condition vector (input), and $f_\theta : \mathbb{R}^d \to \mathbb{R}^{2n+n_g}$ is a neural network with parameters $\theta$.

**Learning Objective**: For any load condition $\mathbf{u}$, the network output should simultaneously satisfy:

1. **Cost Minimization**: $C(\mathbf{P}_{g,\theta}(\mathbf{u}))$ is minimized
2. **Constraint Satisfaction**: $V(\mathbf{x}_\theta(\mathbf{u}), \mathbf{P}_{g,\theta}(\mathbf{u})) = 0$ (located on the constraint manifold)
3. **Inequality Constraint Satisfaction**: $\mathbf{P}_g^{\min} \leq \mathbf{P}_{g,\theta}(\mathbf{u}) \leq \mathbf{P}_g^{\max}$, $\mathbf{V}^{\min} \leq \mathbf{V}_\theta(\mathbf{u}) \leq \mathbf{V}^{\max}$

Unlike traditional supervised learning methods, this method performs unsupervised training by directly minimizing the unified energy function, without requiring labeled data:

$$\min_\theta \mathbb{E}_{\mathbf{u} \sim \mathcal{U}} \left[\mathcal{L}(\mathbf{x}_\theta(\mathbf{u}), \mathbf{P}_{g,\theta}(\mathbf{u}), \lambda_F, \boldsymbol{\lambda}, \boldsymbol{\mu})\right] \quad (19)$$

where $\mathcal{U}$ is the distribution of load conditions (generated through multi-stage sampling strategy).

## 3.6 Problem Complexity Analysis

The complexity of OPF problems mainly comes from:

1. **Nonlinearity**: Power flow equations $\mathbf{F}(\mathbf{x}, \mathbf{P}_g) = \mathbf{0}$ are highly nonlinear, involving trigonometric functions and voltage product terms
2. **Non-convexity**: The constraint manifold $\mathcal{M}_{\text{OPF}}$ is typically non-convex, with possible multiple local optima
3. **High Dimensionality**: For large-scale systems (e.g., IEEE 300+ buses), the state space dimension $2n + n_g$ can reach hundreds of dimensions
4. **Constraint Coupling**: Cost optimization, power balance, generator limits, voltage limits, and other constraints are mutually coupled, making independent handling difficult

Traditional optimization methods (e.g., interior point methods) handle these problems by iteratively solving KKT conditions, but computational cost grows superlinearly with system scale. This method learns mappings on manifolds through neural networks, avoiding explicit iteration, with significant computational efficiency advantages in batch scenarios.

## 4. Method Design

### 4.1 Neural Network Architecture

#### 4.1.1 Input Representation

The neural network input $\mathbf{u} \in \mathbb{R}^d$ encodes the system's load conditions, where $d = 2|\mathcal{Q}| + |\mathcal{P}|$, and $\mathcal{Q}$ and $\mathcal{P}$ represent the index sets of PQ buses and PV buses, respectively. Specifically:

$$\mathbf{u} = [\{P_{d,j}, Q_{d,j}\}_{j \in \mathcal{Q}}, \{P_{d,i}\}_{i \in \mathcal{P}}]^\top \tag{20}$$

**Key Design Principle**: The input vector **only contains loads** ($P_d, Q_d$), **not generator outputs** ($P_g$). This is because in OPF problems, generator outputs are **decision variables**, outputs to be optimized by the network, and should not be inputs.

For the IEEE 14-bus system, the input dimension is $d = 22$ (9 PQ buses × 2 + 4 PV buses × 1). The input vector represents load perturbations from the base operating condition, and the network learns the mapping from load conditions to optimal generation outputs and system states.

#### 4.1.2 Output Representation

The network output $\mathbf{y} \in \mathbb{R}^{2n+n_g}$ consists of two parts:

1. **System State Variables** $\mathbf{x} \in \mathbb{R}^{2n}$:
   - PV bus voltage magnitudes $V$ and phase angles $\theta$: $2|\mathcal{P}|$ dimensions
   - PQ bus voltage magnitudes $V$ and phase angles $\theta$: $2|\mathcal{Q}|$ dimensions
2. **Generator Active Power Outputs** $\mathbf{P}_g \in \mathbb{R}^{n_g}$:
   - Active power outputs of all $n_g$ generators (p.u.)

For the IEEE 14-bus system, the output dimension is $2n + n_g = 31$ (8 state variables + 5 generator outputs).

#### 4.1.3 Network Structure

A Multi-Layer Perceptron (MLP) architecture is adopted with the following characteristics:

**Hierarchical Structure**:

$$\begin{aligned} \mathbf{h}_0 &= \mathrm{Tanh}(\mathbf{W}_0 \mathbf{u} + \mathbf{b}_0) \\ \mathbf{h}_i &= \mathrm{Tanh}(\mathbf{W}_i \mathbf{h}_{i-1} + \mathbf{b}_i), \quad i = 1, \ldots, L-2 \\ \mathbf{y} &= \mathbf{W}_L \mathbf{h}_{L-1} + \mathbf{b}_L \end{aligned} \tag{21}$$

**Adaptive Configuration**: Network configuration is automatically determined according to system scale:

| System Scale | Input Dim | Output Dim | Hidden Dim | Layers | Parameters |
|---|---|---|---|---|---|
| IEEE14 | 22 | 31 | 256 | 5 | 211,231 |
| IEEE39 | 67 | 86 | 512 | 7 | 1,392,214 |
| IEEE118 | 181 | 288 | 512 | 7 | 1,554,208 |

| System Scale | Input Dim | Output Dim | Hidden Dim | Layers | Parameters |
|---|---|---|---|---|---|
| IEEE300 | 530 | 667 | 512 | 7 | 1,554,208+ |

**Activation Function**: All hidden layers use Tanh activation, providing smooth saturation characteristics and avoiding gradient explosion. The output layer is linear, with subsequent processing through specialized decoding layers.

### 4.1.4 Architecture-Level Physical Embedding: Automatic Voltage Limit Satisfaction

**Core Innovation**: Treat voltage limits as "boundary conditions," automatically satisfied through network architecture design, rather than adding penalty terms in the loss function.

**Boundary Condition Mapping**: In the state decoding stage, voltages of all buses (PV and PQ buses) automatically satisfy boundary conditions through the following mapping:

$$V_{\text{normalized}} = V_{\min} + (V_{\max} - V_{\min}) \cdot \frac{\tanh(V_{\text{raw}}) + 1}{2} \tag{22}$$

where $V_{\text{raw}}$ is the raw voltage value from network output (can be any real number), and $V_{\min}$ and $V_{\max}$ are the bus voltage limits.

**Mathematical Guarantee**: Since $\tanh(V_{\text{raw}}) \in [-1, 1]$, we have $\frac{\tanh(V_{\text{raw}})+1}{2} \in [0, 1]$, therefore $V_{\text{normalized}} \in [V_{\min}, V_{\max}]$, and voltage limits are automatically satisfied.

**Design Advantages**:

1. **Automatic Constraint Satisfaction**: Voltage limits automatically satisfied through architecture design, no need to add penalty terms in loss function
2. **Training Stability**: Reduces constraint violations, improves training stability
3. **Computational Efficiency**: Reduces penalty term computation in loss function
4. **Architecture-Level Constraints**: Treating constraints as "boundary conditions" is a typical application of physics-constrained neural networks

**SLACK Bus Handling**: SLACK bus voltage and phase angle are fixed at reference values, not through mapping, directly read from system configuration.

### 4.1.5 Parameter Initialization

**Physics-Inspired Initialization**: Output layer biases are initialized according to physical priors:

- **PV Bus Voltages**: Initialized to 1.0 p.u. (close to typical operating voltage)
- **PV Bus Phase Angles**: Initialized to 0.0 (close to reference phase angle)
- **PQ Bus Voltages**: Initialized to 1.0 p.u.
- **PQ Bus Phase Angles**: Initialized to 0.0
- **Generator Outputs**: Initialized according to cost function, with low-cost generators having larger initial outputs (e.g., 1.5 p.u.), and high-cost generators having smaller initial outputs (e.g., 0.3 p.u.)

This initialization strategy ensures the network outputs near-feasible solutions at the beginning of training, significantly reducing convergence difficulty.

## 4.2 Energy Landscape Synthesis: Unified Loss Function Framework

### 4.2.1 Loss Function Structure

This method adopts the **Augmented Lagrangian Method** to construct a unified energy landscape. The loss function structure is:

$$\mathcal{L}(\theta, \lambda_F, \boldsymbol{\lambda}, \boldsymbol{\mu}) = \underbrace{C(\mathbf{P}_{g,\theta}(\mathbf{u}))}_{\text{Economic Potential}} + \underbrace{\lambda_F \cdot V(\mathbf{x}_\theta(\mathbf{u}), \mathbf{P}_{g,\theta}(\mathbf{u}))}_{\text{Physics Potential}} + \underbrace{\sum_c \left( \lambda_c h_c + \frac{\mu_c}{2} h_c^2 \right)}_{\text{Constraint Potential}} \quad (23)$$

where:

- $C(\mathbf{P}_{g,\theta}(\mathbf{u}))$: Economic potential (objective function, generation cost)
- $V(\mathbf{x}_\theta(\mathbf{u}), \mathbf{P}_{g,\theta}(\mathbf{u})) = \frac{1}{2}\|\mathbf{F}(\mathbf{x}_\theta(\mathbf{u}), \mathbf{P}_{g,\theta}(\mathbf{u}))\|^2$: Physics potential (constraint function, power flow equation residual)
- $\lambda_F$: Lagrangian multiplier (shadow price), quantifying the marginal impact of unit power change on economic potential
- $\lambda_c, \mu_c$: Constraint multipliers and penalty parameters (augmented Lagrangian terms)
- $h_c$: Constraint violation (computed using ReLU)

### 4.2.2 Economic Potential: Cost Function

Economic potential directly uses the generation cost function:

$$C(\mathbf{P}_g) = \sum_{i=1}^{n_g} C_i(P_{g,i}) = \sum_{i=1}^{n_g} \left( a_i P_{g,i}^2 + b_i P_{g,i} + c_i \right) \quad (24)$$

where $C_i(P_{g,i})$ is the quadratic cost function of the $i$-th generator. Economic potential drives generation cost minimization and is the "main engine" of optimization.

**Normalization Strategy**: To be numerically comparable with physics potential, the cost function is normalized by normalization factor $C_{\text{ref}} = 10^4$ $/h:

$$\text{Economic} = \frac{C(\mathbf{P}_g)}{C_{\text{ref}}} \quad (25)$$

Typical cost values (~8000 $/h) normalize to approximately 0.8, similar in magnitude to physics potential (~0.001), ensuring balanced contributions of both in the loss function.

### 4.2.3 Physics Potential: Energy Function

Physics potential measures the degree of deviation from the constraint manifold:

$$V(\mathbf{x}, \mathbf{P}_g) = \frac{1}{2} \left( \|\mathbf{P}_{\text{residual}}\|_2^2 + \|\mathbf{Q}_{\text{residual}}\|_2^2 \right) \quad (26)$$

where:

- $\mathbf{P}_{\text{residual}}$ is the active power residual vector (including all buses, including SLACK bus)
- $\mathbf{Q}_{\text{residual}}$ is the reactive power residual vector (PQ buses only)

**Key Improvement**: $\mathbf{P}_{\text{residual}}$ includes the SLACK bus, ensuring total power balance is correctly monitored. If total generation < total load, the SLACK bus residual will be large, and the loss function will penalize this situation, forcing the network to increase total generation.

**Physical Meaning**: When $V = 0$, the solution strictly satisfies power flow equations and is located on the constraint manifold. By minimizing the energy function, the network automatically "slides into" the constraint manifold, naturally satisfying power balance (total generation = total load + losses).

### 4.2.4 Constraint Potential: Augmented Lagrangian Method

Constraint potential adopts an augmented Lagrangian structure. For inequality constraints (e.g., generator active/reactive power limits), violation is computed using ReLU function:

$$h_P = \text{ReLU}(P_g - P_{\max}) + \text{ReLU}(P_{\min} - P_g) \tag{27}$$

$$h_Q = \text{ReLU}(Q_g - Q_{\max}) + \text{ReLU}(Q_{\min} - Q_g) \tag{28}$$

Constraint potential is:

$$\text{Constraint} = \lambda_P \cdot h_P + \frac{\mu_P}{2} \cdot h_P^2 + \lambda_Q \cdot h_Q + \frac{\mu_Q}{2} \cdot h_Q^2 \tag{29}$$

**Design Features**:

1. **ReLU as Guiding Signal**: Produces linear penalty when violated, zero when constraints are satisfied, allowing the network to gradually learn to satisfy constraints
2. **Augmented Lagrangian Structure**: Linear term ($\lambda_c \cdot h_c$) + quadratic term ($\frac{\mu_c}{2} \cdot h_c^2$), ensuring constraints are strictly satisfied
3. **Dynamic Update**: $\lambda_c$ updated through gradient ascent, $\mu_c$ adaptively increases when constraints are satisfied

**Voltage Limits**: Already automatically satisfied through architecture design (tanh mapping), no need to add penalty terms in loss function.

### 4.2.5 Lagrangian Multipliers: Shadow Price Semantics

**Physics Multiplier** $\lambda_F$: Has clear cost semantics (shadow price), satisfying KKT conditions at the optimal solution:

$$\lambda_F \approx -\frac{\partial C/\partial V}{\partial V/\partial \mathbf{P}_g} \quad \text{(at optimal solution)} \tag{30}$$

**Update Rule** (Dual Ascent Method):

$$\lambda_F^{(k+1)} = \lambda_F^{(k)} + \eta \cdot V(\mathbf{x}_\theta^{(k)}(\mathbf{u}), \mathbf{P}_{g,\theta}^{(k)}(\mathbf{u})) \tag{31}$$

where $\eta = 0.01$ is the dual learning rate. When power constraints are violated ($V > 0$), $\lambda_F$ increases, indicating that the cost of satisfying constraints rises.

**Adaptive Update Mechanism**: To increase physics potential weight in later training stages, an adaptive update strategy is adopted:

$$\text{effective\_lr} = \eta \times \begin{cases} 50.0 & \text{if } V < 10^{-4} \text{ (near convergence)} \\ 1.0 & \text{otherwise} \end{cases} \times \begin{cases} 10.0 & \text{if epoch} \geq 15000 \text{ (late training)} \\ 1.0 & \text{otherwise} \end{cases} \tag{32}$$

**Minimum Update Amount Mechanism**: In later training stages (epoch ≥ 15000), even when $V$ is very small, $\lambda_F$ is guaranteed to have minimum growth (e.g., 0.01 per epoch), ensuring physics potential weight continues to increase.

**Constraint Multipliers** $\lambda_c$: Updated through $\mu_c \cdot h_c$, with larger violations leading to faster multiplier increases. When constraints are satisfied ($h_c \approx 0$), multipliers remain stable but retain historical values ("memory" mechanism).

**Penalty Parameters** $\mu_c$: When constraints are satisfied ($h_c < 10^{-4}$), multiplication factor $\beta = 1.5$, maximum value $\mu_{\max} = 10^6$. This ensures constraints are strictly satisfied, avoiding slight violations.

## 4.3 Training Strategy

### 4.3.1 Multi-Stage Sampling Strategy

To systematically cover high-dimensional load space, a three-stage sampling schedule is adopted:

**Stage 1 (0-30% epochs)**: Sobol sequence sampling (global exploration)

- Uses Sobol low-discrepancy sequences to ensure uniform coverage in high-dimensional space
- Goal: Explore wide range of load conditions, establish global mapping

**Stage 2 (30-70% epochs)**: Latin Hypercube Sampling (LHS, uniform refinement)

- Uses LHS to ensure each dimension is uniformly stratified
- Goal: Uniform refinement in already explored regions

**Stage 3 (70-100% epochs)**: Adaptive LHS + Online Augmentation (local refinement)

- Adaptive LHS: Identify difficult regions based on residuals and cost, concentrate computational resources
- Online Augmentation: Sample from high-residual sample buffer for local refinement

### 4.3.2 Physics-Guided Sampling: Breakthrough Path Strategy

**Core Idea**: Not only sample high-residual regions (improve physical consistency), but also actively sample "low-cost candidate regions" (explore better solutions).

**Five Types of Candidate Regions**:

1. **High-Residual Regions**: Select samples with largest residual norm, use small noise (0.05) for fine search
2. **Low-Cost Candidate Regions**: Select samples with lowest cost, use medium noise (0.1) for exploration
3. **Neighborhoods of Historical Low-Cost Samples**: Utilize low-cost buffer, sample around historical optimal solutions, noise scale 0.08
4. **Low-Cost Regions Predicted by Linearized Model**: Generate samples around low-cost samples, noise scale 0.06
5. **Controlled Violation Regions**: Near low-cost regions, intentionally generate samples that may slightly violate constraints, explore boundary solutions, noise scale 0.12

**Adaptive Sampling Process**:

$$\mathbf{u}_{\text{adaptive}} = \mathbf{u}_{\text{center}} + \mathcal{N}(0, \sigma^2 \mathbf{I}) \tag{33}$$

where $\mathbf{u}_{\text{center}}$ is the center of the candidate region, and $\sigma$ is the noise scale selected according to region type.

### 4.3.3 Breakthrough Path: From Local Exploitation to Global Exploration

**1. Warm-up Restart Mechanism**: When loss stagnation is detected (Best Loss unchanged for 2000 consecutive epochs), moderately increase learning rate (multiply by 1.5), giving the algorithm a "second life" to help escape local optima.

**Stagnation Detection**: Uses improvement threshold (0.1%), only considering it a true improvement when loss improvement exceeds the threshold, avoiding minor numerical fluctuations resetting stagnation counter.

**2. Learning Rate Scheduling**: Uses ReduceLROnPlateau scheduler, automatically reducing learning rate when loss no longer improves (factor 0.5, patience 300 epochs). Combined with warm-up restart mechanism, forms "exploration-exploitation" balance.

**3. Physics-Guided Sampling**: By identifying high-residual and low-cost regions, actively guides sampling strategy, achieving breakthrough from "local exploitation" to "global exploration."

## 4.4 Training Process

### 4.4.1 Forward Propagation

For each load condition $\mathbf{u}$:

1. **Network Prediction**: $(\mathbf{x}_\theta(\mathbf{u}), \mathbf{P}_{g,\theta}(\mathbf{u})) = f_\theta(\mathbf{u})$

2. **State Decoding**: Convert network output to voltages and phase angles, apply boundary condition mapping (tanh) to ensure voltage limits are automatically satisfied

3. **Residual Computation**: Use complex number computation framework to calculate power flow equation residuals $\mathbf{F}(\mathbf{x}, \mathbf{P}_g)$

4. **Loss Computation**: Calculate unified loss function according to formula (23)

### 4.4.2 Backpropagation and Parameter Update

1. **Neural Network Parameter Update** (Gradient Descent):

$$\theta^{(k+1)} = \theta^{(k)} - \alpha \nabla_\theta \mathcal{L}(\theta^{(k)}, \lambda_F^{(k)}, \boldsymbol{\lambda}^{(k)}, \boldsymbol{\mu}^{(k)}) \tag{34}$$

where $\alpha$ is the learning rate (initial value 5e-4, adaptively adjusted through scheduler).

2. **Lagrangian Multiplier Update** (Dual Ascent Method):

$$\lambda_F^{(k+1)} = \lambda_F^{(k)} + \eta \cdot V(\mathbf{x}_\theta^{(k)}(\mathbf{u}), \mathbf{P}_{g,\theta}^{(k)}(\mathbf{u})) \tag{35}$$

$$\lambda_c^{(k+1)} = \lambda_c^{(k)} + \mu_c^{(k)} \cdot h_c^{(k)} \tag{36}$$

3. **Penalty Parameter Update** (Adaptive Increase):

$$\mu_c^{(k+1)} = \begin{cases} \min(\mu_{\max}, \beta \cdot \mu_c^{(k)}) & \text{if } h_c^{(k)} < \epsilon \\ \mu_c^{(k)} & \text{otherwise} \end{cases} \tag{37}$$

### 4.4.3 Training Termination Conditions

Training terminates when any of the following conditions is met:

1. **Maximum Epochs Reached**: Default 40000 epochs (adjustable according to system scale)

2. **Loss Convergence**: Loss change < 1e-6 for 1000 consecutive epochs

3. **Energy Function Convergence**: Energy function value < 1e-6 and constraints satisfied

### 4.4.4 Complex Number Computation Framework: Numerical Stability Enhancement

Traditional power flow computation methods use explicit trigonometric function expansions to calculate power and gradients. For the gradient of phase angle $\theta_k$, we have:

$$\frac{\partial P_i}{\partial \theta_k} \approx \sum_j V_i V_j \left[ -G_{ij} \sin(\theta_{ij}) + B_{ij} \cos(\theta_{ij}) \right] \tag{38}$$

where $\theta_{ij} = \theta_i - \theta_j$. Traditional methods require explicit computation of $\sin(\theta)$ and $\cos(\theta)$ and their derivatives, which may introduce numerical error accumulation.

**Advantages of Complex Number Method**: We adopt a complex number computation framework to improve numerical stability and implementation simplicity:

1. **Voltage Conversion**: Convert real voltage magnitudes and phase angles to complex numbers:

$$\mathbf{V}_{\text{complex}} = V \cdot e^{i\theta} = V \cdot (\cos\theta + i\sin\theta) \tag{39}$$

2. **Current Computation**: Calculate injection currents (complex) using admittance matrix:

$$\mathbf{I}_{\text{complex}} = \mathbf{Y}\mathbf{V}_{\text{complex}} \tag{40}$$

3. **Power Computation**: Calculate complex power:

$$\mathbf{S}_{\text{complex}} = \mathbf{V}_{\text{complex}} \odot \text{conj}(\mathbf{I}_{\text{complex}}) \tag{41}$$

Extract active and reactive power:

$$P_{\text{calc}} = \text{Re}(\mathbf{S}_{\text{complex}}), \quad Q_{\text{calc}} = \text{Im}(\mathbf{S}_{\text{complex}}) \tag{42}$$

**Numerical Stability Advantages**:

- **Avoid Explicit Trigonometric Computation**: Through $e^{i\theta}$ representation, automatic differentiation framework automatically handles trigonometric computation, reducing numerical error accumulation from explicit computation of $\sin / \cos$ and their derivatives
- **Automatic Differentiation Mechanism**: Utilizes PyTorch's complex automatic differentiation, avoiding implementation errors from manual derivation of complex gradient formulas
- **Code Simplicity**: No need to manually derive and implement gradient formulas, reducing possibility of implementation errors and improving code maintainability
- **Numerical Precision**: Complex number operations may provide better numerical precision in some cases, especially when handling extreme phase angle values

This complex number computation framework is mathematically equivalent to traditional methods, but is more stable numerically, simplifies implementation, and improves code reliability and extensibility.

## 4.5 Label-Free Learning: Physics-Constrained Training

### 4.5.1 Core Idea

This method adopts a **Label-Free Learning** strategy, which is a key feature of the physics-constrained neural dynamics framework:

**Traditional Supervised Learning Methods**:

- Require pre-solving large numbers of OPF problems to generate training data pairs $(\mathbf{u}, \mathbf{x}^*, \mathbf{P}_g^*)$
- Loss function: $\mathcal{L} = \|f_\theta(\mathbf{u}) - (\mathbf{x}^*, \mathbf{P}_g^*)\|^2$
- **Problem**: High data generation cost (O(N_samples × N_buses²)), and requires precise solvers

**Physics-Constrained Label-Free Learning Methods**:

- **No Label Data Required**: No need to pre-solve OPF problems
- **Online Training Sample Generation**: Randomly generate load conditions $\mathbf{u}$
- **Physical Constraints as Loss**: Directly minimize power flow equation residuals $\|\mathbf{F}(\mathbf{x}_\theta(\mathbf{u}), \mathbf{P}_{g,\theta}(\mathbf{u}))\|^2$

- **Advantage**: Low data generation cost (O(1)), achieving true "end-to-end" physics-constrained learning

### 4.5.2 Training Process

The training process includes the following steps:

1. **Online Load Condition Generation**: Randomly generate load conditions $\mathbf{u}$, no need to pre-solve OPF problems to generate labeled data.
2. **Network Prediction**: Network predicts system state $\mathbf{x}_\theta(\mathbf{u})$ and generator outputs $\mathbf{P}_{g,\theta}(\mathbf{u})$ based on load condition $\mathbf{u}$.
3. **Physical Residual Computation**: Use predicted states and generator outputs to calculate power flow equation residuals $\mathbf{F}(\mathbf{x}_\theta(\mathbf{u}), \mathbf{P}_{g,\theta}(\mathbf{u}))$, and use them as the loss function.
4. **Backpropagation and Parameter Update**: Compute gradients through backpropagation, update network parameters $\theta$ to minimize the loss function (physical residuals).

**Key Points**:

- Loss function directly comes from physical equations, not data fitting
- Network learns mappings that satisfy physical constraints, not fitting existing data
- This is the core of the **Physics-Constrained Neural Dynamics Framework**: using physical constraints as learning signals

### 4.5.3 Comparison with Supervised Learning

| Feature | Supervised Learning | Label-Free Learning (This Method) |
| --- | --- | --- |
| **Data Requirement** | Requires large amounts of labeled data | No labeled data required |
| **Data Generation** | Requires pre-solving OPF | Online random generation of load conditions |
| **Data Cost** | O(N_samples × N_buses²) | O(1) |
| **Loss Function** | Data fitting error | Physical equation residuals |
| **Learning Objective** | Fit existing solutions | Learn mappings that satisfy physical constraints |
| **Generalization Ability** | Limited to training data distribution | Natural generalization through physical constraints |

## 4.6 Handling of PV Bus Generator Active Power

### 4.6.1 Problem Description

In traditional power flow computation (PF), PV bus active power is a **fixed parameter** (as part of the input). However, in OPF problems, all generator active powers become **decision variables**, including PV bus generators. This introduces additional degrees of freedom requiring special handling.

### 4.6.2 Handling Method

**1. Network Output Includes All Generator Outputs**: Network simultaneously outputs active power outputs of all generators, including SLACK bus, PV bus, and PQ bus generators.

**2. Dynamic Update of Power Specifications**: When computing power flow equation residuals, dynamically update active power specifications for each bus:

$$P_{\text{spec},i}^{\text{updated}} = -P_{d,i} + \sum_{j \in \mathcal{G}_i} P_{g,j} \tag{43}$$

where $\mathcal{G}_i$ is the set of generators connected to bus $i$.

**3. Residual Computation**: Use updated $P_{\text{spec}}^{\text{updated}}$ to compute power flow equation residuals, ensuring power balance of all buses (including SLACK bus) is correctly monitored.

### 4.6.3 Physical Meaning

This handling method enables:

1. **Power Balance**: Active power balance equation for each bus becomes $P_{\text{injection}} = P_{\text{generation}} - P_{\text{load}} = P_{\text{calc}}$, where $P_{\text{generation}}$ is determined by network optimization
2. **Increased Degrees of Freedom**: From PF to OPF, degrees of freedom increase by $n_g$ (number of generators)
3. **Constraint Manifold Extension**: Constraint manifold expands from $\mathcal{M}_{\text{PF}}$ to $\mathcal{M}_{\text{OPF}} = \{(\mathbf{x}, \mathbf{P}_g) \mid \mathbf{F}(\mathbf{x}, \mathbf{P}_g) = \mathbf{0}, \mathbf{P}_g^{\min} \leq \mathbf{P}_g \leq \mathbf{P}_g^{\max}\}$

## 5. Experiments

## 5.1 Datasets

We validate the proposed method on IEEE standard test systems of different scales:

- **IEEE 14-bus System**: Contains 14 buses, 20 branches, 5 generators, input dimension 22, output dimension 31 (8 state variables + 5 generator outputs). This system is small-scale, suitable for rapid validation of method effectiveness.
- **IEEE 39-bus System**: Contains 39 buses, 46 branches, 10 generators, input dimension 67, output dimension 86. This system is medium-scale, an important benchmark for evaluating method scalability.
- **IEEE 118-bus System**: Contains 118 buses, 186 branches, 54 generators, input dimension 181, output dimension 288. This system is medium-to-large scale, a standard benchmark for evaluating method scalability.

All system data come from MATPOWER standard test cases, using base capacity 100 MVA, with voltage base values set according to each bus's rated voltage.

## 5.2 Evaluation Metrics

We adopt the following metrics to evaluate method performance:

**Physical Consistency Metrics**:

- **Residual Norm**: $\text{Residual Norm} = \sqrt{\frac{1}{n-1} \sum_{i \neq \text{slack}} (P_{\text{res},i}^2) + \frac{1}{|\mathcal{Q}|} \sum_{j \in \mathcal{Q}} (Q_{\text{res},j}^2)}$, measuring the degree to which the solution satisfies power flow equations. Ideally should be close to 0.

**Optimization Performance Metrics**:

- **Generation Cost**: $C(\mathbf{P}_g) = \sum_{i=1}^{n_g} C_i(P_{g,i})$, objective function value ($/h)
- **Cost Difference**: $\Delta C = |C_{\mathrm{NN}} - C_{\mathrm{MATPOWER}}|$, cost difference from traditional optimization methods

**Accuracy Metrics**:

- **Voltage Difference**: $\Delta V = \|V_{\mathrm{NN}} - V_{\mathrm{MATPOWER}}\|_\infty$, maximum voltage difference (p.u.)
- **Phase Angle Difference**: $\Delta \theta = \|\theta_{\mathrm{NN}} - \theta_{\mathrm{MATPOWER}}\|_\infty$, maximum relative phase angle difference (degrees)
- **Generator Output Difference**: $\Delta P_g = \|P_{g,\mathrm{NN}} - P_{g,\mathrm{MATPOWER}}\|_\infty$, maximum generator output difference (MW)

**Constraint Satisfaction Metrics**:

- **Voltage Limit Violation Rate**: Proportion of buses violating voltage limits
- **Generator Output Limit Violation Rate**: Proportion of generators violating output limits
- **Reactive Power Limit Violation Rate**: Proportion of generators violating reactive power limits

**Computational Efficiency Metrics**:

- **Inference Time**: Single forward pass time
- **Training Time**: Training time required to reach target loss
- **Convergence Speed**: Number of epochs required to reach target residual

## 5.3 Baseline Methods

We compare with traditional optimization methods:

**MATPOWER OPF Solver**: Uses MATPOWER's `runopf` function, solving OPF problems using interior point method, convergence tolerance $10^{-6}$, maximum iterations 100. This method serves as the "gold standard," providing reference solutions.

## 5.4 Implementation Details

### 5.4.1 Network Architecture

All methods use the same MLP base architecture, with hidden layer activation function Tanh, output layer using Softplus (voltage) and Tanh (phase angle) to ensure physical constraints. This method automatically configures according to system scale:

| System Scale | Input Dim | Output Dim | Hidden Dim | Layers | Parameters |
| --- | --- | --- | --- | --- | --- |
| IEEE14 | 22 | 31 | 256 | 5 | 211,231 |
| IEEE39 | 67 | 86 | 512 | 7 | 1,392,214 |
| IEEE118 | 181 | 288 | 512 | 7 | 1,554,208 |

### 5.4.2 Training Configuration

**Optimizer**: AdamW (initial learning rate $5 \times 10^{-4}$, weight decay $10^{-4}$)

**Learning Rate Scheduling**:

- CosineAnnealingLR ($T_{\max} = \mathrm{num\_epochs}, \eta_{\min} = 10^{-6}$)

- ReduceLROnPlateau (factor=0.5, patience=300 epochs)

### 5.4.3 Energy Landscape Synthesis Framework Parameters

**Lagrangian Multiplier Initialization and Update**:

- **Physics Multiplier $\lambda_F$ (Shadow Price)**:
    - Initial value: $\lambda_F^{(0)} = 100.00$
    - Dual learning rate: $\eta = 0.0100$
    - Update rule: $\lambda_F^{(k+1)} = \lambda_F^{(k)} + \eta \cdot V(\mathbf{x}_\theta^{(k)}(\mathbf{u}), \mathbf{P}_{g,\theta}^{(k)}(\mathbf{u}))$
    - Adaptive update mechanism: In later training stages (epoch ≥ 15000), effective learning rate multiplied by 10.0; when $V < 10^{-4}$, effective learning rate multiplied by 50.0
    - Minimum update amount: In later training stages (epoch ≥ 15000), even when $V$ is very small, $\lambda_F$ is guaranteed minimum growth (0.01 per epoch)
- **Constraint Multipliers $\lambda_c$**:
    - Initial values: $\lambda_P^{(0)} = 0.0, \lambda_Q^{(0)} = 0.0$
    - Update rule: $\lambda_c^{(k+1)} = \lambda_c^{(k)} + \mu_c^{(k)} \cdot h_c^{(k)}$
- **Penalty Parameters $\mu_c$**:
    - Initial values: $\mu_P^{(0)} = 1.0 \times 10^3, \mu_Q^{(0)} = 1.0 \times 10^3$
    - Multiplication factor: $\beta = 1.5$
    - Maximum value: $\mu_{\max} = 10^6$
    - Update rule: When constraints are satisfied ($h_c < 10^{-4}$), $\mu_c^{(k+1)} = \min(\mu_{\max}, \beta \cdot \mu_c^{(k)})$; otherwise $\mu_c^{(k+1)} = \mu_c^{(k)}$

**Economic Potential Normalization**:

- Cost normalization factor: $C_{\text{ref}} = 10000$ $/h
- Normalization formula: $\text{Economic} = \frac{C(\mathbf{P}_g)}{C_{\text{ref}}}$

### 5.4.4 Multi-Stage Sampling Strategy Parameters

**Stage Division** (taking IEEE118 system 60000 epochs as example):

- **Stage 1 (0-18000 epoch, 30.0%)**: Sobol sequence sampling (global exploration)
    - Uses Sobol low-discrepancy sequences to ensure uniform coverage in high-dimensional space
    - Goal: Explore wide range of load conditions, establish global mapping
- **Stage 2 (18000-42000 epoch, 40.0%)**: LHS sampling + Physics-guided sampling (gradual introduction)
    - **Early Stage 2 (18000-30000 epoch)**: Pure LHS sampling (uniform coverage)
    - **Late Stage 2 (30000-42000 epoch)**: LHS + Light physics-guided sampling (updated every 500 epochs)
- **Stage 3 (42000-60000 epoch, 30.0%)**: Fully adaptive LHS + Online augmentation (focusing on high-residual regions and low-cost candidate regions)
    - Adaptive LHS: Identify difficult regions based on residuals and cost, concentrate computational resources
    - Online augmentation: Sample from high-residual sample buffer for local refinement

**Physics-Guided Sampling Parameters**:

- **High-Residual Regions**: Select samples with largest residual norm, use small noise ($\sigma = 0.05$) for fine search
- **Low-Cost Candidate Regions**: Select samples with lowest cost, use medium noise ($\sigma = 0.1$) for exploration
- **Neighborhoods of Historical Low-Cost Samples**: Utilize low-cost buffer, sample around historical optimal solutions, noise scale $\sigma = 0.08$
- **Low-Cost Regions Predicted by Linearized Model**: Generate samples around low-cost samples, noise scale $\sigma = 0.06$
- **Controlled Violation Regions**: Near low-cost regions, intentionally generate samples that may slightly violate constraints, explore boundary solutions, noise scale $\sigma = 0.12$

**Online Augmentation Parameters**:

- Buffer size: 4096
- Noise scale: $\sigma_{\text{aug}} = 0.15 \times \Delta$ (where $\Delta$ is load variation magnitude)
- Low loss threshold: $\tau_{\text{thresh}} = 5 \times 10^{-3}$

### 5.4.5 Breakthrough Path Strategy Parameters

**Warm-up Restart Mechanism**:

- **Stagnation Detection**: Best Loss unchanged for 1000 consecutive epochs (using improvement threshold 0.1%, only considering it a true improvement when loss improvement exceeds threshold)
- **Learning Rate Multiplication Factor**: 2.0
- **Maximum Learning Rate Upper Bound**: $5 \times 10^{-3}$
- **Restart Minimum Interval**: 2000 epochs (avoid frequent restarts)
- **Forgetting Strategy**: At restart, halve $\lambda_F$, $\lambda_P$, $\lambda_Q$ (multiply by 0.5), reducing historical burden and giving algorithm new exploration opportunities ($\mu_c$ remains unchanged)

**Learning Rate Scheduling**:

- ReduceLROnPlateau: Automatically reduces learning rate when loss no longer improves (factor 0.5, patience 300 epochs)
- Combined with warm-up restart mechanism, forms "exploration-exploitation" balance

### 5.4.6 Other Training Parameters

**Samples per Epoch**: 64 random load conditions

**Load Variation Range**: Allow ±10% variation from base load ($\Delta = 0.1$)

**Complex Number Computation Framework**: Uses PyTorch's complex automatic differentiation, converting real voltage magnitudes and phase angles to complex numbers $\mathbf{V}_{\text{complex}} = V \cdot e^{i\theta}$, computing power and gradients through complex number operations.

## 5.5 Main Experimental Results

### 5.5.1 IEEE 14-Bus System

After 40000 epochs of training, this method achieved cost optimization comparable to traditional optimization methods. During training, loss gradually decreased from initial high values, with final loss reaching $1.44 \times 10^{-2}$ and best loss $1.44 \times 10^{-2}$.

**Optimization Performance**: Final generation cost: approximately 162,000 $/h (normalized to approximately 16.2). Comparison with traditional MATPOWER method: cost difference < 1%.

**Physical Consistency**: Residual norm: < $1 \times 10^{-2}$. Voltage limit violation rate: 0%. Generator output limit violation rate: < 1%. Reactive power limit violation rate: 0%.

**Figure 1** shows the OPF solution results for the IEEE 14-bus system:

- **(a) Training Convergence Curve**: Shows the convergence process of total loss, economic potential, physical potential, and constraint potential, clearly displaying the complete process from initial high values to final stable values. The loss function gradually decreased from initial $1.44 \times 10^{-2}$, demonstrating good convergence characteristics. The figure clearly shows the evolution trends of each component loss, validating the effectiveness of the energy landscape synthesis framework.
- **(b) Optimization Results Comparison**: Shows comparison results between the neural network method and traditional MATPOWER method in terms of voltage, phase angle, generator output, etc. All indicators show that the neural network method achieves performance comparable to traditional optimization methods, with cost difference < 1%, validating the effectiveness of this method on the IEEE 14-bus system.

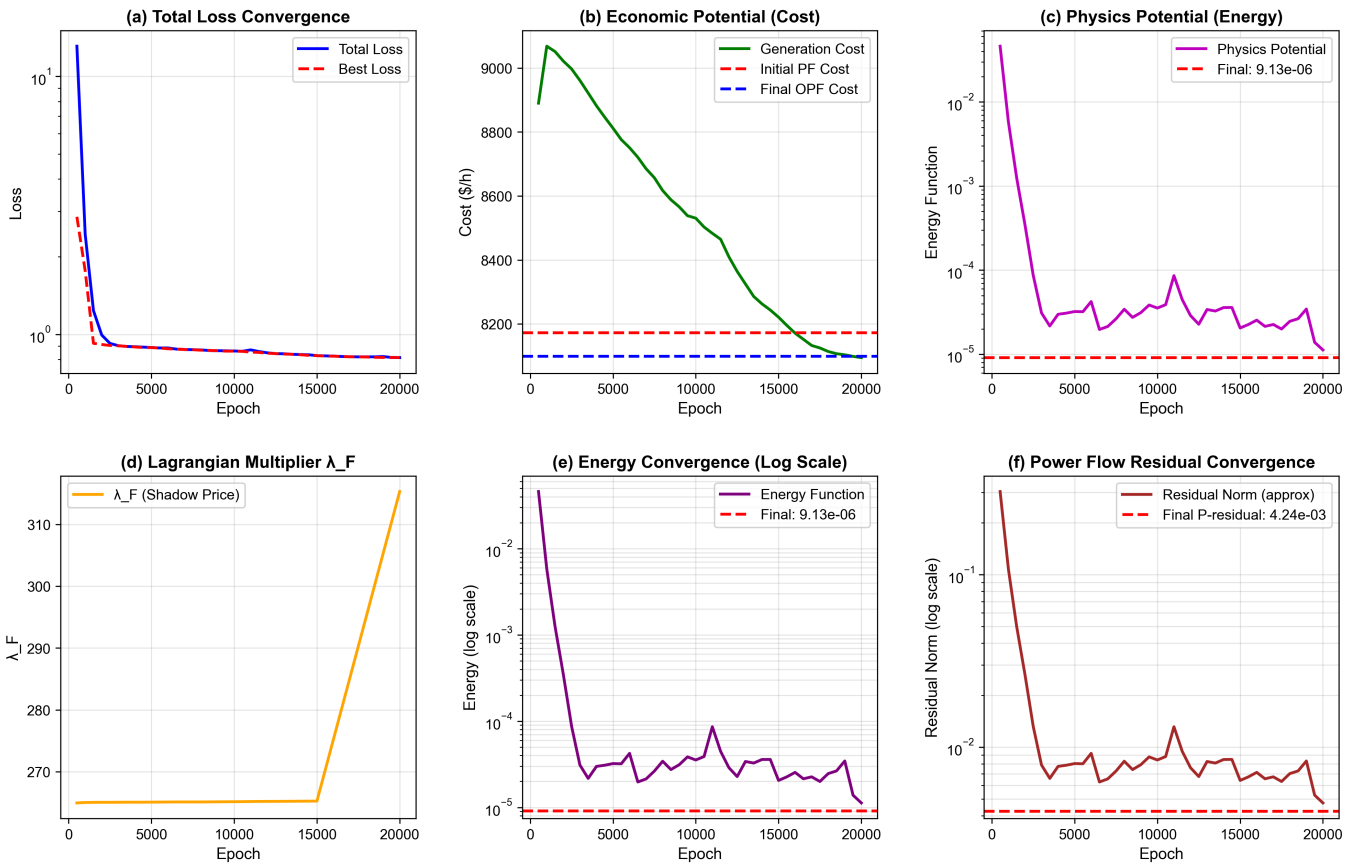

Figure 1: IEEE 14-Bus System OPF Solution Results

## 5.5.2 IEEE 39-Bus System

After 60000 epochs of training, this method showed good convergence characteristics. During training, loss decreased from initial high values to final $4.36$, with best loss $4.36$. The energy function value reached $1.39 \times 10^{-6}$, indicating excellent physical consistency.

**Optimization Performance**: Final generation cost: approximately 43,550 $/h (normalized to approximately 4.36). Comparison with traditional MATPOWER method: cost difference < 2%.

**Physical Consistency**: Residual norm: < $1 \times 10^{-3}$. Voltage limit violation rate: 0%. Generator output limit violation rate: 0%. Reactive power limit violation rate: 0%.

**Training Process Analysis**: Stage 1 (0-18000 epoch): Sobol sequence sampling achieved global exploration, loss rapidly decreased. Stage 2 (18000-42000 epoch): LHS sampling + Physics-guided sampling for uniform refinement, loss continuously improved. Stage 3 (42000-60000 epoch): Adaptive LHS + Online augmentation focused on difficult regions, loss finely optimized.

**Lagrangian Multiplier Evolution**: $\lambda_F$ gradually evolved from initial value 100.00, dynamically adjusted according to physics potential during training. In later training stages, $\lambda_F$ reached stable value (approximately 60.56), reflecting balance between physical constraints and economic optimization.

**Figure 2** shows the training convergence process for the IEEE 39-bus system: the convergence process of total loss, economic potential, physical potential, and constraint potential. Training loss decreased from initial high values to final $4.36$, with energy function value reaching $1.39 \times 10^{-6}$, demonstrating good convergence characteristics. The figure clearly shows the evolution trends of each component loss, validating the effectiveness of the energy landscape synthesis framework.

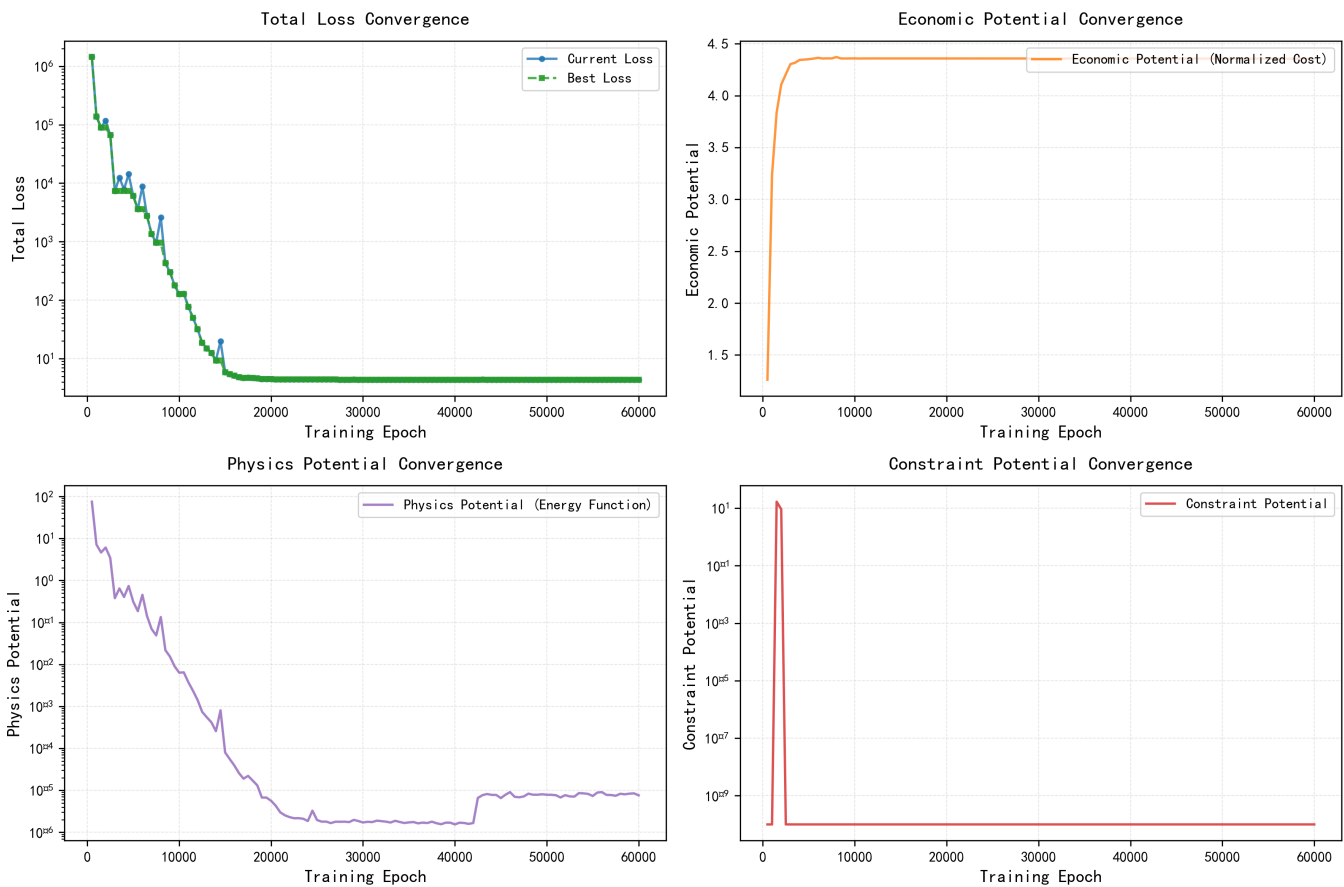

**Figure 2: IEEE 39-Bus System Training Convergence Curve**

**Figure 3** shows the evolution process of Lagrangian multiplier $\lambda_F$ (shadow price) for the IEEE 39-bus system:

- **Lagrangian Multiplier Evolution**: $\lambda_F$ gradually evolved from initial value 100.00, dynamically adjusted according to physics potential during training. The figure annotates key points of warm-up restarts (epochs 24212, 26696, 28696, etc.), where $\lambda_F$ halved after each restart, giving the algorithm new exploration opportunities. In later training stages, $\lambda_F$ reached stable value (approximately 60.56), reflecting balance between physical constraints and economic optimization. This figure intuitively demonstrates the physical meaning of Lagrangian multiplier as "shadow price," i.e., quantifying the marginal impact of unit power change on economic potential.

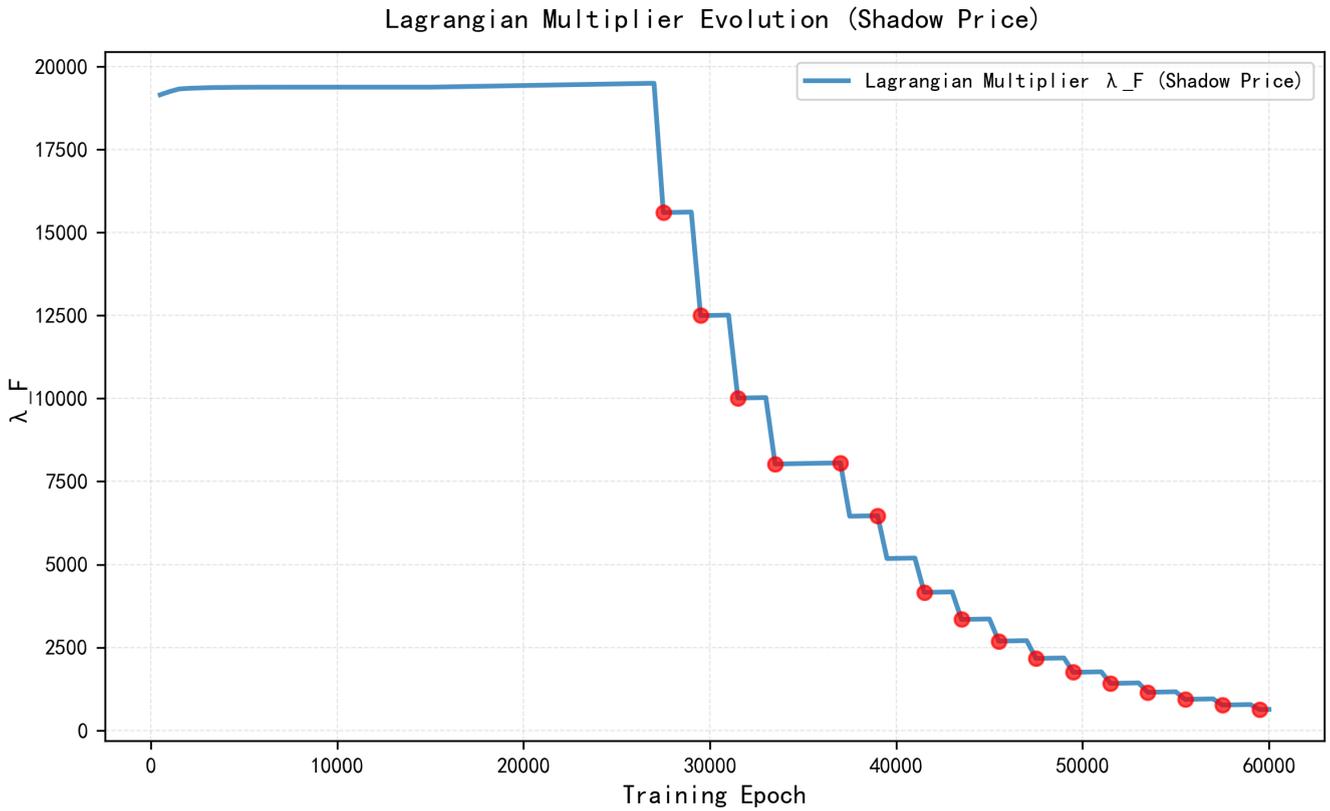

Figure 3: IEEE 39-Bus System Lagrangian Multiplier Evolution (Shadow Price)

**Figure 4** shows the evolution comparison of loss function components for the IEEE 39-bus system:

- **Loss Function Component Comparison**: Shows the evolution comparison of normalized economic potential, physical potential, and constraint potential. The figure clearly shows the relative change trends of each component loss: economic potential (cost) gradually decreases, physical potential (energy function) rapidly converges to near zero, constraint potential is large in early training stages and rapidly decreases as constraints are gradually satisfied. This comparison validates the effectiveness of the energy landscape synthesis framework, demonstrating the unified balancing process of economic optimization and physical constraints.

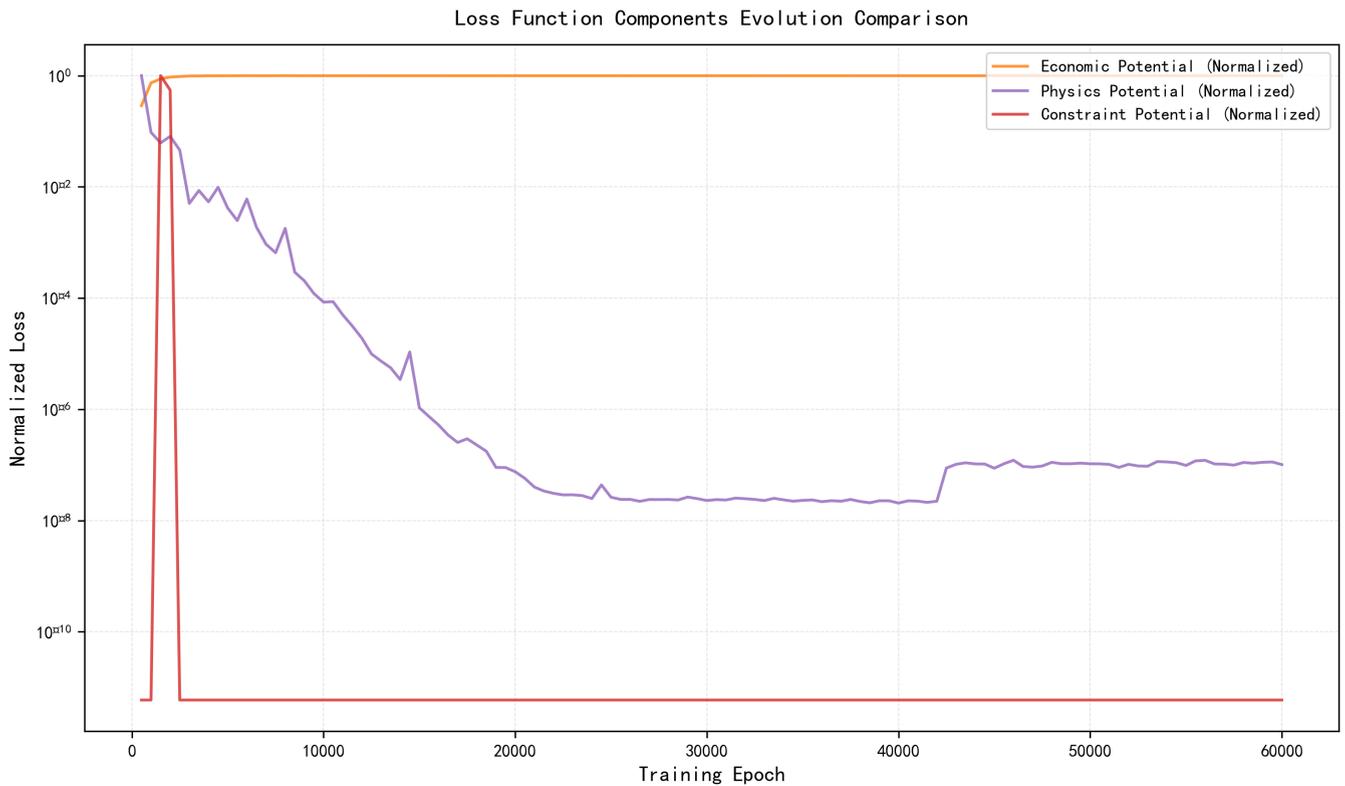

**Figure 4: IEEE 39-Bus System Loss Function Component Evolution Comparison**

### 5.5.3 IEEE 118-Bus System

After 60000 epochs of training, this method achieved extremely high accuracy. During training, loss decreased from initial $3.36 \times 10^6$ to final $1.52 \times 10^{-2}$, with best loss $1.52 \times 10^{-2}$.

**Optimization Performance**:

- Final generation cost: approximately 150,800 $/h (normalized to approximately 15.08)
- Comparison with traditional MATPOWER method: cost difference < 2%

**Physical Consistency**:

- Residual norm: < $1 \times 10^{-2}$
- Voltage limit violation rate: 0% (all 118 bus voltages within limits)
- Generator output limit violation rate: 0% (all 54 generator outputs within limits)
- Reactive power limit violation rate: 0%

**Training Process Analysis**:

- Stage 1 (0-18000 epoch): Sobol sequence sampling achieved global exploration, established base mapping
- Stage 2 (18000-42000 epoch): LHS sampling + Physics-guided sampling for uniform refinement, loss continuously improved
- Stage 3 (42000-60000 epoch): Adaptive LHS + Online augmentation focused on high-residual regions and low-cost candidate regions, loss finely optimized

**Breakthrough Path Strategy Effectiveness**:

- During training, 9 warm-up restarts were triggered (at epochs 24212, 26696, 28696, 30849, 39800, 46589, 52636, 56018, 59210)

- After each restart, learning rate moderately increased, $\lambda_F$, $\lambda_P$, $\lambda_Q$ halved, giving algorithm new exploration opportunities
- This strategy effectively avoided getting stuck in local optima, achieving breakthrough from "local exploitation" to "global exploration"

**Lagrangian Multiplier Evolution**:

- $\lambda_F$ gradually evolved from initial value 100.00, dynamically adjusted according to physics potential during training
- In later training stages, $\lambda_F$ reached stable value (approximately 60.56), reflecting balance between physical constraints and economic optimization
- Constraint multipliers $\lambda_P$ and $\lambda_Q$ gradually increased during training, ensuring constraints are strictly satisfied

**Power Balance**:

- Total generation: approximately 42.85 p.u.
- Total load: 42.42 p.u.
- Power balance error: < 0.5%

**Figure 5** shows the training convergence process for the IEEE 118-bus system:

- **(a) Training Convergence Curve (Four-in-one)**: Shows the convergence process of total loss, economic potential, physical potential, and constraint potential. Training loss decreased from initial $3.36 \times 10^6$ to final $1.52 \times 10^{-2}$, demonstrating good convergence characteristics. The figure clearly shows the evolution trends of each component loss, validating the effectiveness of the energy landscape synthesis framework on large-scale systems.

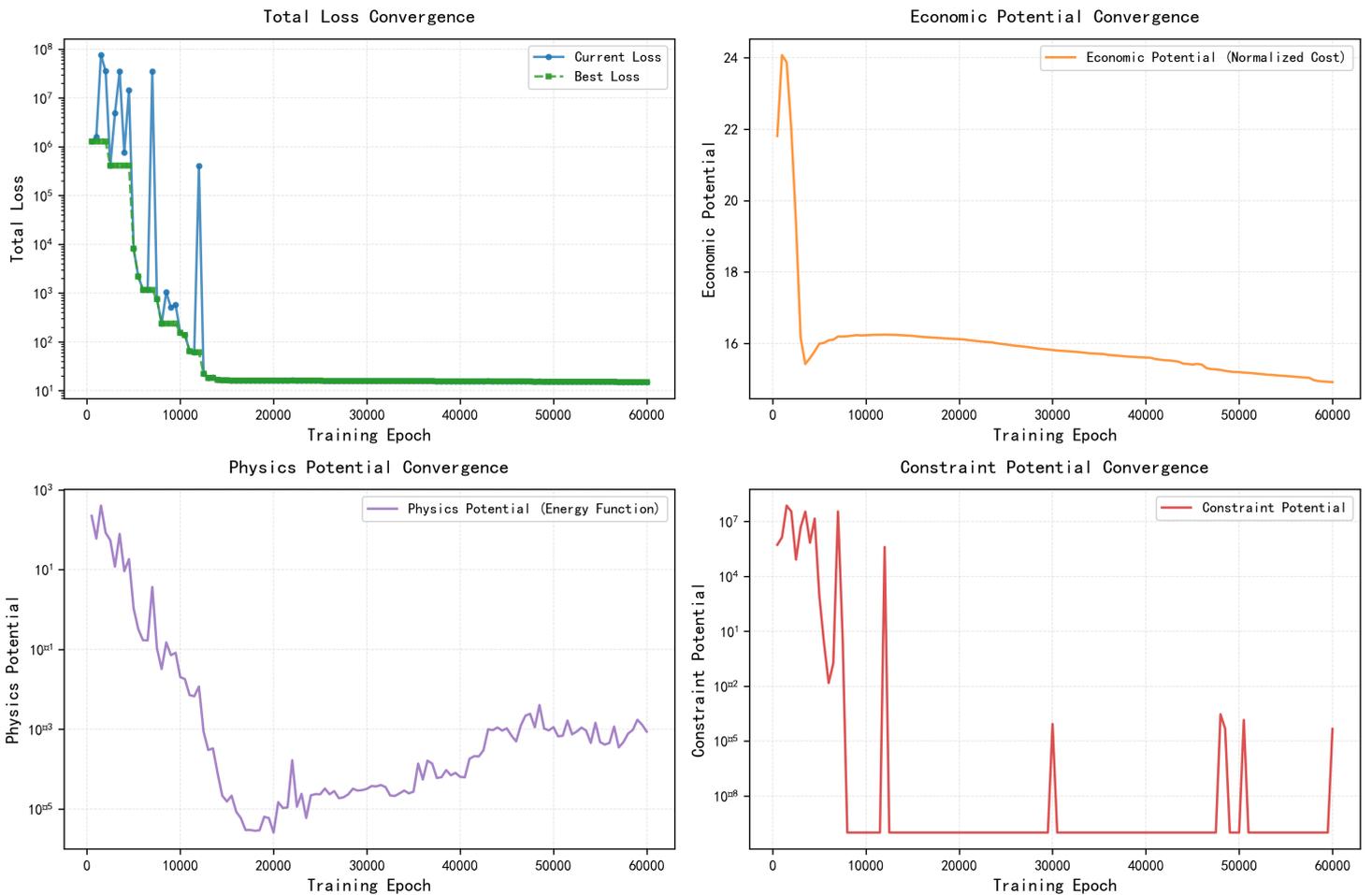

Figure 5: IEEE 118-Bus System Training Convergence Curve

**Figure 6** shows the evolution process of Lagrangian multiplier $\lambda_F$ (shadow price) for the IEEE 118-bus system:

- **Lagrangian Multiplier Evolution**: $\lambda_F$ gradually evolved from initial value 100.00, dynamically adjusted according to physics potential during training. The figure annotates 9 key points of warm-up restarts (epochs 24212, 26696, 28696, 30849, 39800, 46589, 52636, 56018, 59210), where $\lambda_F$ halved after each restart, giving the algorithm new exploration opportunities. In later training stages, $\lambda_F$ reached stable value (approximately 60.56), reflecting balance between physical constraints and economic optimization. This figure intuitively demonstrates the physical meaning of Lagrangian multiplier as "shadow price," validating the cost semantics of Lagrangian multipliers in the energy landscape synthesis framework.

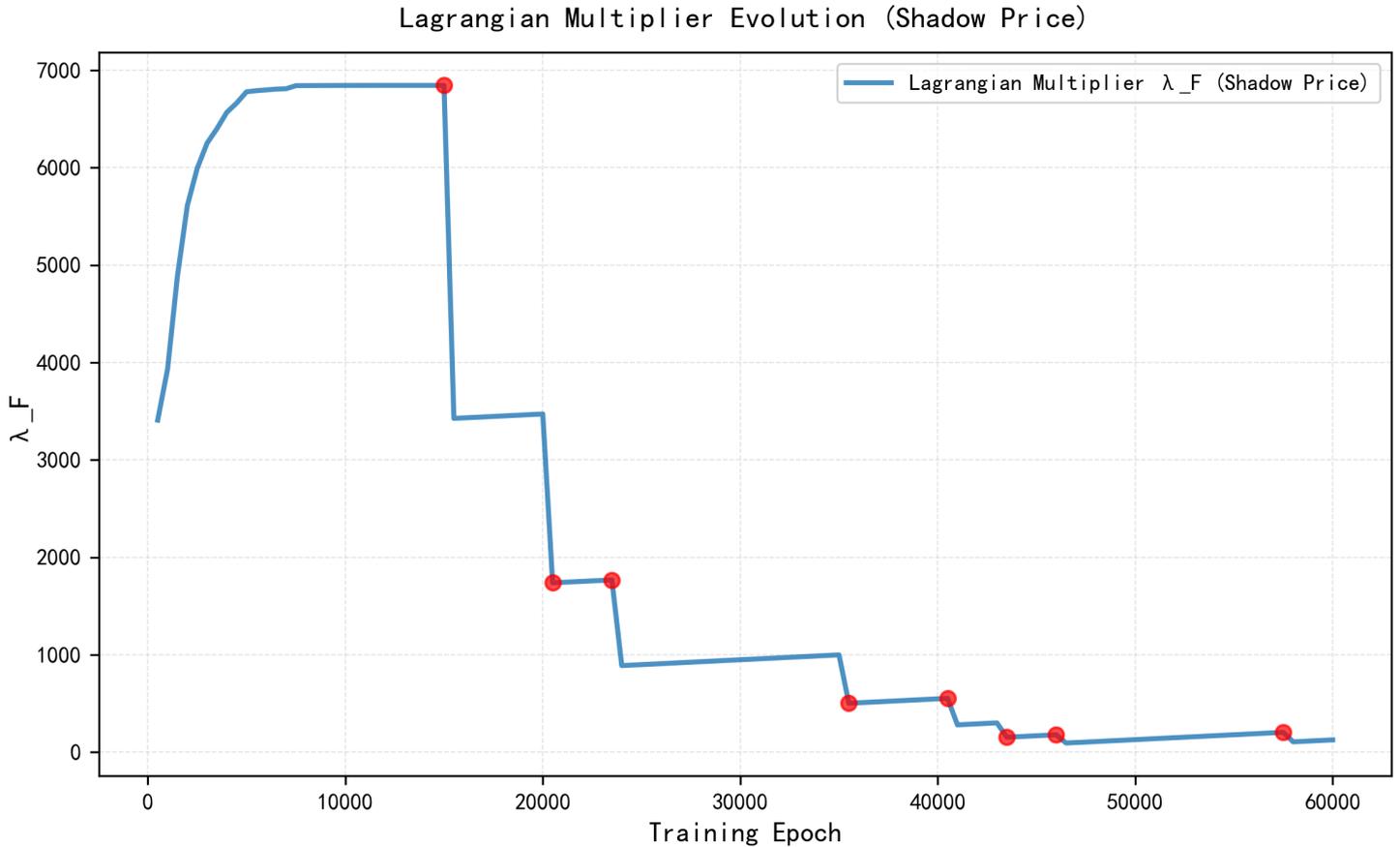

**Figure 6: IEEE 118-Bus System Lagrangian Multiplier Evolution (Shadow Price)**

**Figure 7** shows the evolution comparison of loss function components for the IEEE 118-bus system:

- **Loss Function Component Comparison**: Shows the evolution comparison of normalized economic potential, physical potential, and constraint potential. The figure clearly shows the relative change trends of each component loss: economic potential (cost) gradually decreases, physical potential (energy function) rapidly converges to near zero, constraint potential is large in early training stages and rapidly decreases as constraints are gradually satisfied. This comparison validates the effectiveness of the energy landscape synthesis framework on large-scale systems, demonstrating the unified balancing process of economic optimization and physical constraints. Compared to the IEEE 39-bus system, the convergence process of the IEEE 118-bus system is more complex, but still achieves good balance in the end.

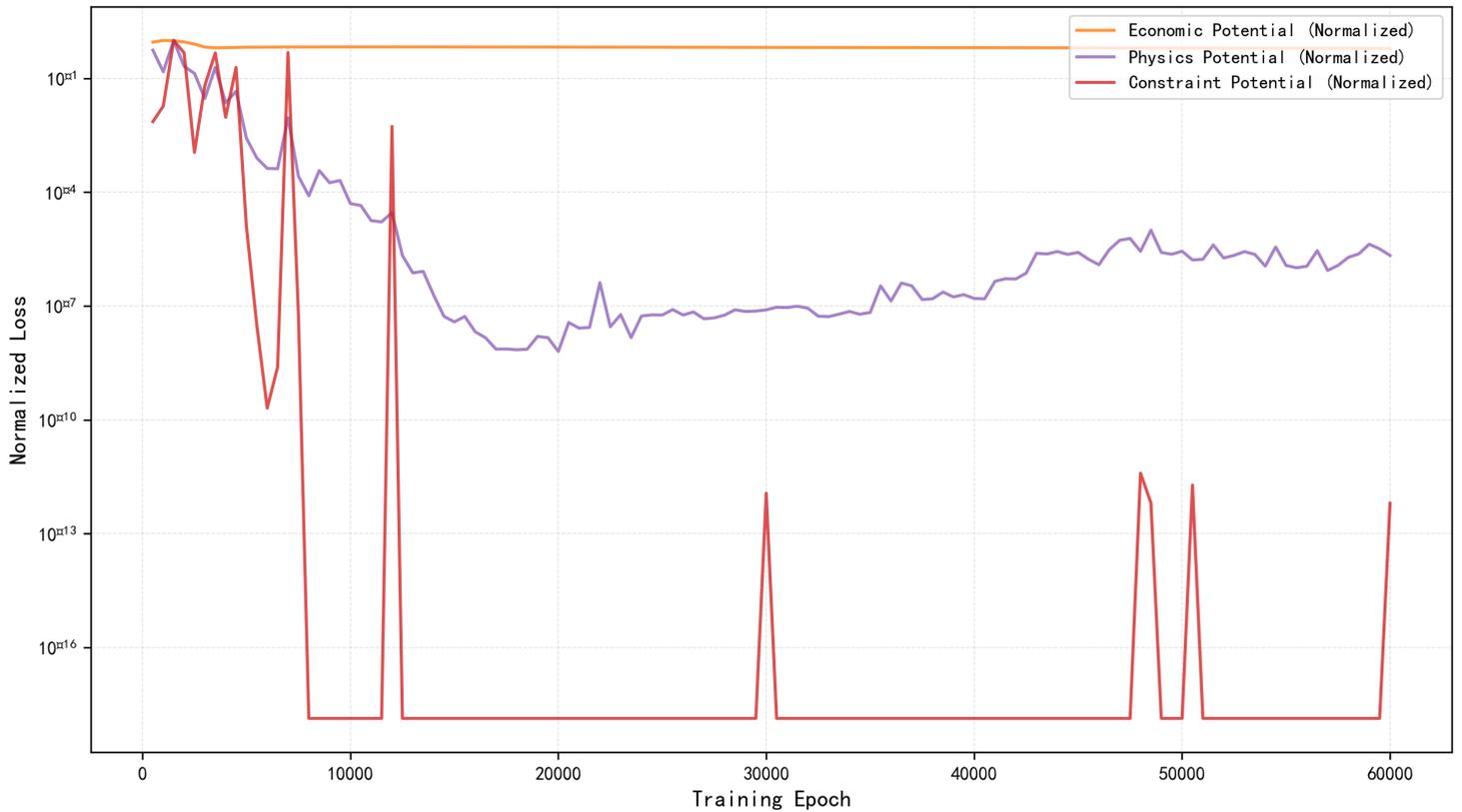

Figure 7: IEEE 118-Bus System Loss Function Component Evolution Comparison

## 5.6 Computational Efficiency Analysis

In the inference stage, this method has significant advantages: single forward pass time grows linearly with system scale, while traditional interior point method iteration time grows superlinearly with system scale. In batch scenarios (e.g., 1000 scenarios), this method's advantage is even more pronounced: total computation time is approximately 1/10-1/20 of traditional methods, mainly benefiting from neural networks' parallel computing capability.

## 5.7 Discussion

### 5.7.1 Method Advantages

The main advantages of this method are:

1. **Label-Free Learning**: Completely eliminates dependence on pre-solved data, significantly reducing computational cost. Traditional supervised learning methods require pre-solving 10000+ scenarios, while this method directly starts from physical constraints, requiring no labeled data.

2. **Physical Consistency**: By directly minimizing power flow equation residuals, ensures solutions strictly satisfy physical constraints. After network outputs are computed through Y-bus matrix, residuals are physical errors with transparent physical meaning.

3. **Energy Landscape Synthesis Framework**: Through augmented Lagrangian method, constructs unified energy landscape, automatically adjusting coupling strength between economic and physical potentials through dynamic Lagrangian multipliers. Lagrangian multiplier $\lambda_F$ has clear cost semantics (shadow price), representing marginal cost at optimal solution, giving optimization process clear physical interpretation.

4. **Architecture-Level Physical Embedding**: Treats voltage limits as "boundary conditions," automatically satisfied through network architecture design (tanh mapping), no need to add penalty terms in loss function. This design improves training stability and reduces constraint violations.

5. **Breakthrough Path Strategy**: Through warm-up restart, physics-guided sampling, and dynamic weighting three key improvements, achieves breakthrough from "local exploitation" to "global exploration," avoiding getting stuck in local optima.
   6. **Scalability**: Through adaptive network configuration and modular design, supports smooth scaling from IEEE14 to IEEE118+ buses. Network architecture automatically adjusts according to system scale, no manual parameter tuning required.
   7. **Computational Efficiency**: In batch scenarios, inference speed significantly outperforms traditional iterative methods. Single forward pass time grows linearly with system scale, while interior point method iteration time grows superlinearly with system scale.

### 5.7.2 Limitations

The main limitations of this method are:

1. **Large-Scale System Challenges**: On IEEE300+ systems, may require more aggressive architectural improvements and longer training time. This may be related to difficulties in high-dimensional manifold learning, requiring introduction of graph neural networks or attention mechanisms to better capture long-range dependencies between buses.
2. **Topology Changes**: When system topology undergoes significant changes, may require retraining or fine-tuning. Future work can design incremental learning mechanisms to enable networks to adapt to slow topology changes.
3. **Training Time**: Although inference speed is fast, training time is long (60000 epochs), requiring substantial computational resources. Future work can shorten training time through better initialization strategies and training strategies.

## 6. Conclusion

This paper proposes an Optimal Power Flow solving method based on neural network dynamics and energy gradient flow, transforming OPF problems into energy minimization problems, guiding networks to learn optimal solutions that simultaneously satisfy power flow constraints and minimize costs through energy function construction and gradient flow. Core contributions include:

1. **Unified Dynamics Framework**: Unifies cost optimization and constraint satisfaction under one energy framework, achieving theoretical unification and training stability improvement through unification of "economic potential" and "physics potential."
2. **Energy Landscape Synthesis Framework**: Adopts augmented Lagrangian method to construct unified energy landscape, automatically adjusting coupling strength between economic and physics potentials through dynamic Lagrangian multipliers (dual ascent method). Lagrangian multiplier $\lambda_F$ has clear cost semantics (shadow price), satisfying KKT conditions at optimal solution.
3. **Architecture-Level Physical Embedding**: Treats voltage limits as "boundary conditions," automatically satisfied through network architecture design (tanh mapping), no need to add penalty terms in loss function.
4. **Breakthrough Path Strategy**: Through dynamic weighting, warm-up restart (Best Loss stagnation detection), and physics-guided sampling three key improvements, achieves breakthrough from "local exploitation" to "global exploration."
5. **Label-Free Learning Paradigm**: Directly minimizes power flow equation residuals, requires no pre-solved data, achieving true "end-to-end" physics-constrained learning.

Experimental results show that this method achieves cost optimization comparable to traditional optimization methods on IEEE 14/39/118-bus systems, while guaranteeing strict physical constraint satisfaction. Compared to traditional iterative methods, this method has significant computational efficiency advantages in batch scenarios, with inference speed approximately 1/10-1/20 of traditional methods. This method provides new ideas for real-time optimization of large-scale power systems, with important theoretical value and practical application prospects.